\documentclass[a4paper,10pt]{article}

\usepackage{graphicx}

\usepackage{graphicx,wrapfig}
\usepackage{caption}
\usepackage{subcaption}
\usepackage{dsfont}
\usepackage{amsmath}
\usepackage{amssymb}
\usepackage{amsthm}
\usepackage{amsfonts}
\usepackage{mathtools}
\usepackage{algorithm}
\usepackage{algorithmic}
\usepackage{hyperref}
\usepackage{cleveref}
\usepackage{bm}
\usepackage{stmaryrd}
\usepackage{xcolor}
\usepackage[export]{adjustbox}
\usepackage{stackengine}
\usepackage{booktabs}
\usepackage{makecell}
\usepackage{tikz}
\usepackage{pgfplots}
\usetikzlibrary{spy,calc}

\usepackage[colorinlistoftodos,prependcaption,textsize=tiny,textwidth=0.2\textwidth]{todonotes}

\usepackage[export]{adjustbox}

\newtheorem{proposition}{Proposition}
\newtheorem{assumption}{Assumption}

\DeclareMathOperator*{\argmin}{arg\,min}
\DeclareMathOperator*{\argmax}{arg\,max}

\DeclareMathAlphabet\mathbfcal{OMS}{cmsy}{b}{n}

\newcommand{\MSE}{\mathrm{MSE}}
\newcommand{\MMSE}{ {\scriptscriptstyle\mathrm{MMSE}} }
\newcommand{\prox}{\mathrm{prox}}

\newcommand{\dist}{\mathrm{dist}}

\newcommand{\Id}{\mathbf{\mathrm{Id}}}

\newcommand{\Fc}{\mathcal{F}}
\newcommand{\Pc}{\mathcal{P}}
\newcommand{\Ac}{\mathcal{A}}
\newcommand{\Tc}{\mathcal{T}}

\newcommand{\Nc}{\mathcal{N}}
\newcommand{\Dc}{\mathcal{D}}

\newcommand{\R}{\mathbb{R}}
\newcommand{\N}{\mathbb{N}}

\newcommand{\C}{\mathbb{C}}
\newcommand{\K}{\mathbb{K}}

\newcommand{\E}{\mathbb{E}}
\renewcommand{\P}{\mathbb{P}}

\newcommand{\eqdef}{\stackrel{\mathrm{def}}{=}}

\newcommand{\Ab}{\mathbf{A}}

\newcommand{\hb}{\mathbf{h}}
\newcommand{\xb}{\mathbf{x}}

\newcommand{\yb}{\mathbf{y}}

\newcommand{\bb}{\mathbf{b}}

\newcommand{\Az}{A}

\newcommand{\wz}{w}
\newcommand{\kz}{k}

\newcommand{\xz}{x}

\newcommand{\yz}{y}
\newcommand{\zz}{z}
\newcommand{\bz}{b}

\newcommand{\ssb}{\mathbf{s}}

\newcommand{\muz}{\mu}
\newcommand{\alphab}{\bm{\alpha}}
\newcommand{\xib}{\boldsymbol{\xi}}
\newcommand{\thetab}{\boldsymbol{\theta}}

\newcommand{\omegab}{\boldsymbol{\omega}}
\newcommand{\lambdab}{\boldsymbol{\lambda}}

\newcommand{\thetaz}{\theta}

\newcommand{\red}[1]{\textcolor{red}{#1}}
\newcommand{\blue}[1]{\textcolor{blue}{#1}}
\newcommand{\rev}[1]{#1}

\setstackgap{S}{-1.5pt}

\makeatletter
\newcommand{\plotwithzoom}[1]{%
\begin{tikzpicture}[spy using outlines={circle,size=.52\textwidth, magnification=3, connect spies}] %
    \node[inner sep=0pt] (img) 
    at (-.375\textwidth,0) { \includegraphics[width=\textwidth]{#1} }; %
    \begin{scope}[x={($ (img.south east) - (img.south west) $ )},y={( $ (img.north west) - (img.south west)$ )}, shift={(img.south west)}] %
        \node (spy1n) at    (0.26,0.71) {};
        \coordinate (spy1nto) at (.73,0.27); %
        \spy [white,thick] on (spy1n) in node at (spy1nto);%
    \end{scope}%
\end{tikzpicture} %
}
\makeatother

\usepackage[a4paper,margin=1in,footskip=0.25in]{geometry}

\begin{document}

\title{Training Adaptive Reconstruction Networks for Blind Inverse Problems}

\author{Alban Gossard\textsuperscript{1} \\ \footnotesize \href{mailto:alban.paul.gossard@gmail.com}{alban.paul.gossard@gmail.com}
\and Pierre Weiss\textsuperscript{2} \\ \footnotesize \href{mailto:pierre.weiss@cnrs.fr}{pierre.weiss@cnrs.fr}
\and \footnotesize \textsuperscript{1} Institut de Mathématiques de Toulouse (IMT), Université de Toulouse, CNRS, France
\and \footnotesize \parbox{0.8\linewidth}{\textsuperscript{2} Institut de Recherche en Informatique de Toulouse (IRIT) et Centre de Biologie Intégrative (CBI), équipe MCD, CNRS, Université de Toulouse, France}
}
\date{\today}

\maketitle

\begin{abstract}
    Neural networks allow solving many ill-posed inverse problems with unprecedented performance. Physics informed approaches already progressively replace carefully hand-crafted reconstruction algorithms in real applications. However, these networks suffer from a major defect: when trained on a given forward operator, they do not generalize well to a different one. The aim of this paper is twofold. First, we show through various applications that training the network with a family of forward operators allows solving the adaptivity problem without compromising the reconstruction quality significantly.
    Second, we illustrate that this training procedure allows tackling challenging blind inverse problems.
    \rev{Our experiments include partial Fourier sampling problems arising in magnetic resonance imaging (MRI) with sensitivity estimation and off-resonance effects, computerized tomography (CT) with a tilted geometry  and image deblurring with Fresnel diffraction kernels.}
\end{abstract}


\def\imagewidth{0.3\textwidth}
\def\tabwidth{0.45\textwidth}
\def\tabwidthfamily{0.8\textwidth}
\def\tabwidthfamilyct{0.4\textwidth}


\section{Introduction}

The primary contribution of this paper is the design of model-based neural networks to solve \emph{families} of blind inverse problems. Many sensing devices like cameras, Magnetic Resonance Imaging (MRI) or Computerized Tomography (CT) systems measure a signal $\xz\in \K^N$ through a linear operator $\Az(\thetaz) \in \K^{M\times N}$ with $\K= \R$ or $\K=\C$.
The parameter $\thetaz\in \R^P$ characterizes the sensing operator. For instance, it can encode the point spread function in image deblurring, the projection angles in CT or the Fourier sampling locations \rev{and coil sensitivities} in MRI.
This leads to measurements of the form:
\begin{equation}\label{eq:forward_model}
\yz = \Pc(\Az(\thetaz)\xz),
\end{equation}
where $\Pc:\K^M\to \K^M$ is a perturbation (e.g. additive Gaussian noise, quantization).
\rev{A model-based inverse problem solver constructs} an estimate $\hat \xz$ of $\xz$ from $\yz$ and $\Az(\thetaz)$.
If the parameter $\thetaz$ is unknown, then we speak of a blind inverse problem.
In this paper, we focus on neural network based reconstructions. We consider mappings of the form:
\begin{alignat}{3}
\Nc: \quad & \R^D \times \K^{M\times N} \times \K^M  && \to  \quad \R^N \nonumber \\
     &      (\wz,\Az,\yz)              && \mapsto  \quad \Nc(\wz,\Az,\yz). \label{eq:reconmapping}
\end{alignat}
Given a weight $\wz\in \R^D$, a forward operator $\Az$ and a measurement vector $\yz$, the network $\Nc$ outputs an estimate $\hat \xz=\Nc(\wz,\Az,\yz)$.
\rev{The network depends on the operator $\Az$ since it typically consists in alternating an inversion of $\Az$ followed by a regularization with a neural network.} \rev{For a fixed} forward operator $\Az(\thetaz_0)$, the traditional procedure to train the network consists in optimizing the weights $\wz$ by minimizing the \rev{risk:
\begin{equation}\label{eq:standard_training_procedure}
    \inf_{\wz \in \R^D} R(\wz) \quad \mbox{with} \quad R(\wz) \eqdef \E_{\xb,\yb}\left[ \|\Nc(\wz, \Az(\thetaz_0), \yb) - \xb \|_2^2\right].
\end{equation}
In this equation, $\E_{\xb,\yb}$ indicates the expectation with respect to the random vector $(\xb,\yb)$. 
The random vector $\yb$ is generated using the forward model~\eqref{eq:forward_model}. 
Ideally $\xb$ should be a random vector describing the distribution of images to reconstruct. 
Unfortunately, it is usually unknown and approximated by a discrete probability measure of the form $\frac{1}{I}\sum_{i=1}^I \delta_{\xz_i}$, where $(\xz_i)_{1\leq i \leq I}$ is a collection of training images. We then speak of empirical risk minimization.}
In words, we wish the reconstruction mapping $\Nc(\wz,\cdot,\cdot)$ to output images close in average to the true underlying signals.
In this paper, we explore a seemingly minor variation of this principle by solving: 
\begin{equation}\label{eq:main_idea_of_the paper}
    \inf_{\wz \in \R^D} E(\wz) \quad \mbox{with} \quad E(\wz) \eqdef \E_{\xb,\yb,\thetab}\left[ \|\Nc(\wz, \Az(\thetab), \yb) - \xb \|_2^2\right],
\end{equation}
where the expectation is also taken with respect to the parameter $\thetab$ considered as a random vector. 
That is, we train our reconstruction mapping on a \emph{distribution of operators}.
\rev{While this idea is quite natural and most likely implemented already in a few methods, we believe that this paper is the first to address a systematic study of its performance.}
The main motivation for this modification is twofold. First, we want to address a lack of adaptivity for the standard training procedure. Second, we want to use the resulting reconstruction mapping to solve blind inverse problems. 
\rev{Let us discuss these two points in more depth.}

\paragraph{\rev{Training mismatch issue}}

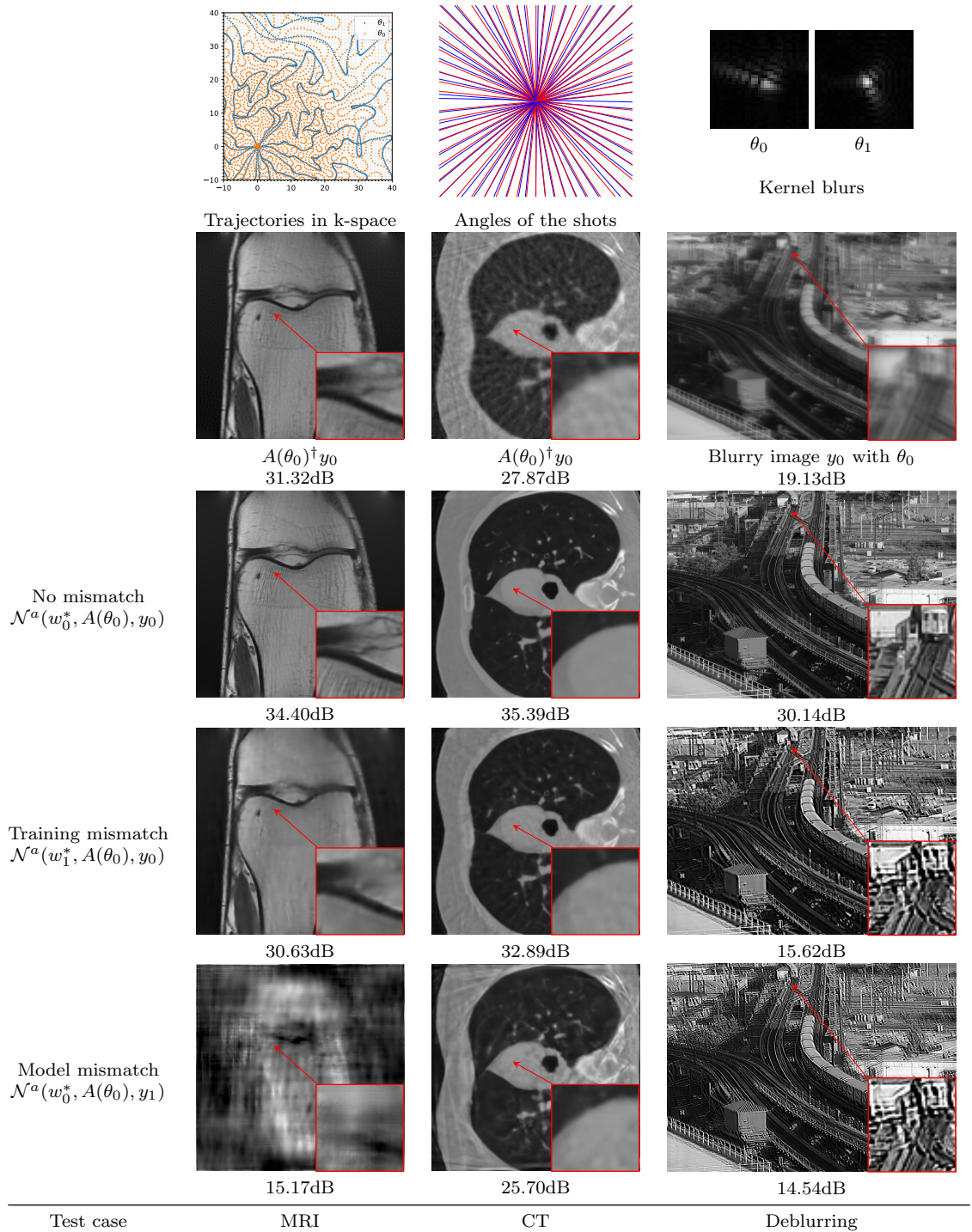
\begin{figure}[t!]
    \def\subfigwidth{0.2\textwidth}
    \def\subfigwidthblur{0.278\textwidth}

    \def\trimzoomleftmri{.3}
    \def\trimzoomlowermri{.6}
    \def\trimzoomrightmri{.55}
    \def\trimzoomuppermri{.25}

    \def\trimzoomleftct{.3}
    \def\trimzoomrightct{.53}
    \def\trimzoomlowerct{.48}
    \def\trimzoomupperct{.35}

    \def\trimzoomleftblur{.38}
    \def\trimzoomlowerblur{.85}
    \def\trimzoomrightblur{.54}
    \def\trimzoomupperblur{.03}
    \centering
    \footnotesize
    \begin{tabular}{@{}ccccc@{}}
         &
        \makecell{\includegraphics[width=\subfigwidth]{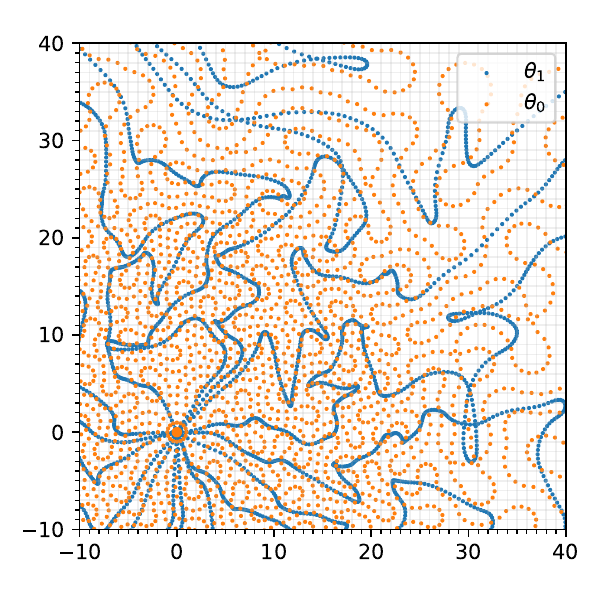}\\ Trajectories in k-space} &
        \makecell{\includegraphics[width=\subfigwidth]{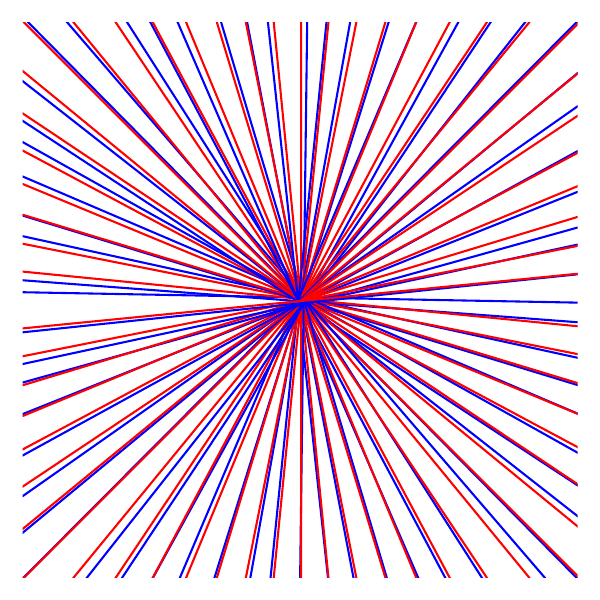}\\ Angles of the shots} &
        \makecell{
            \makecell{\includegraphics[width=1.5cm]{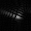}\\$\thetaz_0$\\$\quad$}
            \makecell{\includegraphics[width=1.5cm]{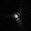}\\$\thetaz_1$\\$\quad$}
            \\ Kernel blurs
        } \\

         &
        \makecell{
        \begin{tikzpicture}
            \node[inner sep=0pt] (img) at (0,0) {
                \includegraphics[width=\subfigwidth]{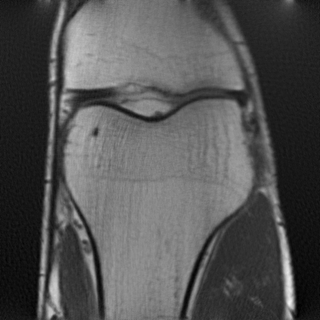}%
                \llap{\adjincludegraphics[width=0.083\textwidth,trim={{\trimzoomleftmri\width} {\trimzoomlowermri\height} {\trimzoomrightmri\width} {\trimzoomuppermri\height}},clip,cfbox=red 0.5pt 0pt]{f_tilde_pseudoinv_Nlines=16_idimg=20_theta0_fixed.png}}
            };
            \draw [stealth-,red] (-0.4,0.35) -- (0.27,-0.25);
        \end{tikzpicture}
        \\ $\Az(\thetaz_0)^\dagger \yz_0$\\ $31.32$dB} &
        \makecell{
        \begin{tikzpicture}
            \node[inner sep=0pt] (img) at (0,0) {
                \includegraphics[width=\subfigwidth]{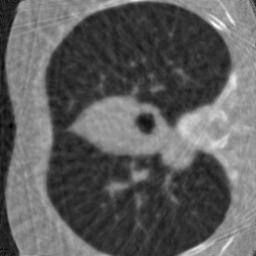}%
                \llap{\adjincludegraphics[width=0.083\textwidth,trim={{\trimzoomleftct\width} {\trimzoomlowerct\height} {\trimzoomrightct\width} {\trimzoomupperct\height}},clip,cfbox=red 0.5pt 0pt]{fig1c_inv27.87.png}}
            };
            \draw [stealth-,red] (-0.35,0.1) -- (0.27,-0.25);
        \end{tikzpicture}
        \\ $\Az(\thetaz_0)^\dagger \yz_0$\\ $27.87$dB} &
        \makecell{
        \begin{tikzpicture}
            \node[inner sep=0pt] (img) at (0,0) {
                \includegraphics[width=\subfigwidthblur]{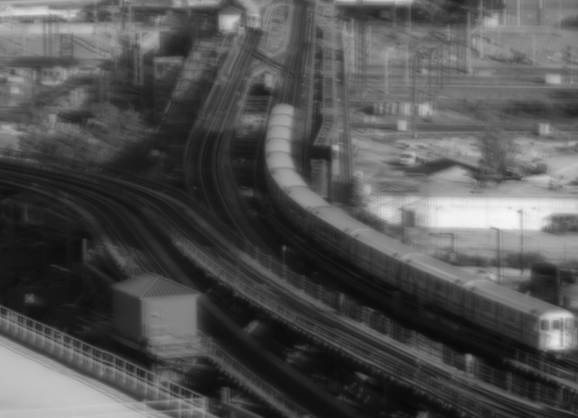}%
                \llap{\adjincludegraphics[width=0.083\textwidth,trim={{\trimzoomleftblur\width} {\trimzoomlowerblur\height} {\trimzoomrightblur\width} {\trimzoomupperblur\height}},clip,cfbox=red 0.5pt 0pt]{y_coco_idimg=10_seed=7_noiselevel=0.0001_theta0_fixed.png}}
            };
            \draw [stealth-,red] (-0.3,1.3) -- (0.85,-0.15);
        \end{tikzpicture}
        \\ Blurry image $\yz_0$ with $\thetaz_0$\\ $19.13$dB} \\

        {\makecell{No mismatch \\ $\Nc^a(\wz_0^*, \Az(\thetaz_0), \yz_0)$}} &
        \makecell{
        \begin{tikzpicture}
            \node[inner sep=0pt] (img) at (0,0) {
                \includegraphics[width=\subfigwidth]{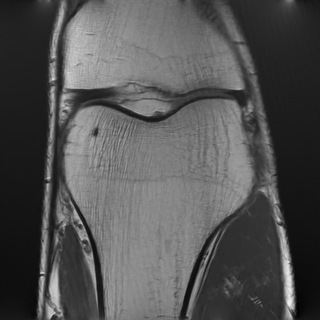}%
                \llap{\adjincludegraphics[width=0.083\textwidth,trim={{\trimzoomleftmri\width} {\trimzoomlowermri\height} {\trimzoomrightmri\width} {\trimzoomuppermri\height}},clip,cfbox=red 0.5pt 0pt]{f_tilde_Nlines=16_idimg=20_theta0_fixed.png}}
            };
            \draw [stealth-,red] (-0.4,0.35) -- (0.27,-0.25);
        \end{tikzpicture}
        \\ $34.40$dB} &
        \makecell{
        \begin{tikzpicture}
            \node[inner sep=0pt] (img) at (0,0) {
                \includegraphics[width=\subfigwidth]{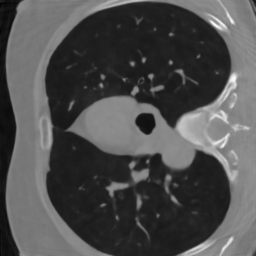}%
                \llap{\adjincludegraphics[width=0.083\textwidth,trim={{\trimzoomleftct\width} {\trimzoomlowerct\height} {\trimzoomrightct\width} {\trimzoomupperct\height}},clip,cfbox=red 0.5pt 0pt]{fig1c_b_35.39.png}}
            };
            \draw [stealth-,red] (-0.35,0.1) -- (0.27,-0.25);
        \end{tikzpicture}
        \\ $35.39$dB} &
        \makecell{
        \begin{tikzpicture}
            \node[inner sep=0pt] (img) at (0,0) {
                \includegraphics[width=\subfigwidthblur]{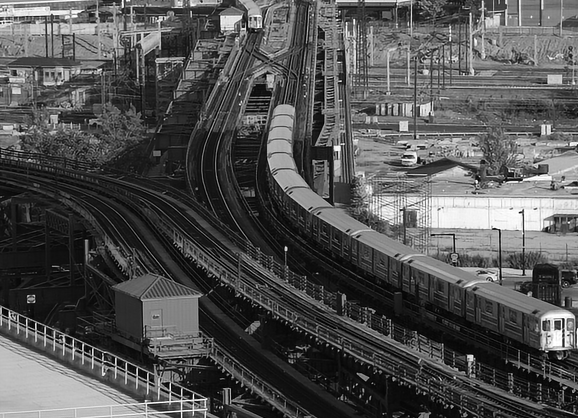}%
                \llap{\adjincludegraphics[width=0.083\textwidth,trim={{\trimzoomleftblur\width} {\trimzoomlowerblur\height} {\trimzoomrightblur\width} {\trimzoomupperblur\height}},clip,cfbox=red 0.5pt 0pt]{f_tilde_coco_idimg=10_seed=7_noiselevel=0.0001_theta0_fixed.png}}
            };
            \draw [stealth-,red] (-0.3,1.3) -- (0.85,-0.15);
        \end{tikzpicture}
        \\ $30.14$dB} \\

        {\makecell{Training mismatch\\ $\Nc^a(\wz_1^*, \Az(\thetaz_0), \yz_0)$}} &
        \makecell{
        \begin{tikzpicture}
            \node[inner sep=0pt] (img) at (0,0) {
                \includegraphics[width=\subfigwidth]{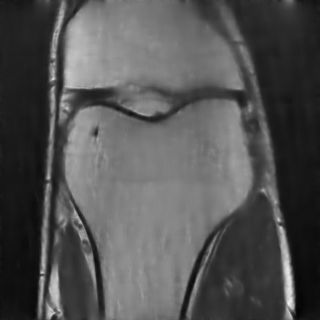}%
                \llap{\adjincludegraphics[width=0.083\textwidth,trim={{\trimzoomleftmri\width} {\trimzoomlowermri\height} {\trimzoomrightmri\width} {\trimzoomuppermri\height}},clip,cfbox=red 0.5pt 0pt]{f_tilde_Nlines=16_idimg=20_nonblind_fixed.png}}
            };
            \draw [stealth-,red] (-0.4,0.35) -- (0.27,-0.25);
        \end{tikzpicture}
        \\ $30.63$dB} &
        \makecell{
        \begin{tikzpicture}
            \node[inner sep=0pt] (img) at (0,0) {
                \includegraphics[width=\subfigwidth]{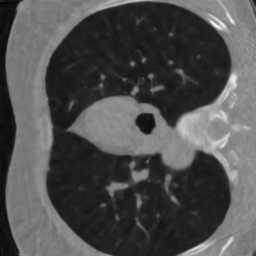}%
                \llap{\adjincludegraphics[width=0.083\textwidth,trim={{\trimzoomleftct\width} {\trimzoomlowerct\height} {\trimzoomrightct\width} {\trimzoomupperct\height}},clip,cfbox=red 0.5pt 0pt]{fig1c_c_32.89.png}}
            };
            \draw [stealth-,red] (-0.35,0.1) -- (0.27,-0.25);
        \end{tikzpicture}
        \\ $32.89$dB} &
        \makecell{
        \begin{tikzpicture}
            \node[inner sep=0pt] (img) at (0,0) {
                \includegraphics[width=\subfigwidthblur]{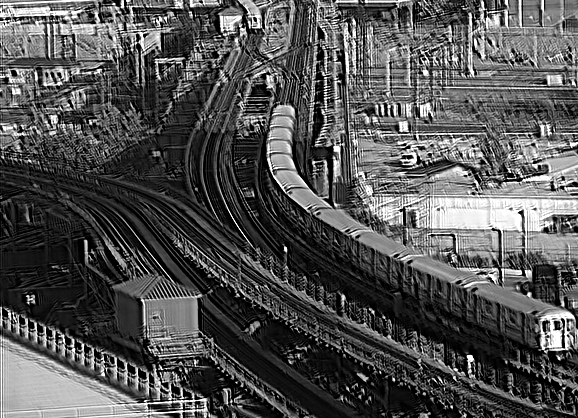}%
                \llap{\adjincludegraphics[width=0.083\textwidth,trim={{\trimzoomleftblur\width} {\trimzoomlowerblur\height} {\trimzoomrightblur\width} {\trimzoomupperblur\height}},clip,cfbox=red 0.5pt 0pt]{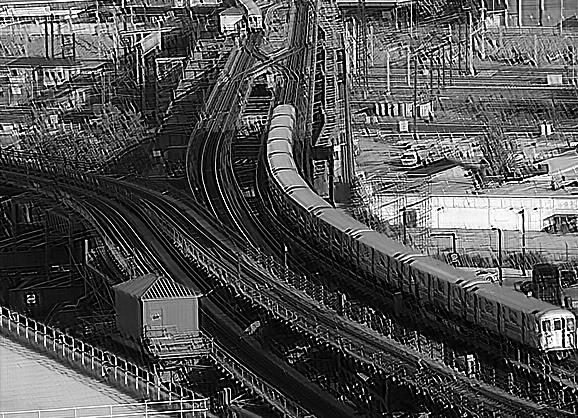}}
            };
            \draw [stealth-,red] (-0.3,1.3) -- (0.85,-0.15);
        \end{tikzpicture}
        \\ $15.62$dB} \\

        {\makecell{Model mismatch\\ $\Nc^a(\wz_0^*, \Az(\thetaz_0), \yz_1)$}} &
        \makecell{
        \begin{tikzpicture}
            \node[inner sep=0pt] (img) at (0,0) {
                \includegraphics[width=\subfigwidth]{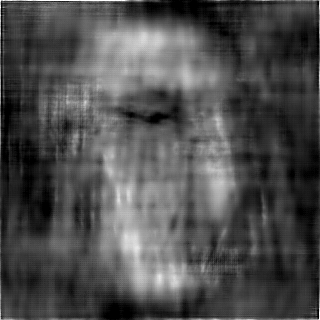}%
                \llap{\adjincludegraphics[width=0.083\textwidth,trim={{\trimzoomleftmri\width} {\trimzoomlowermri\height} {\trimzoomrightmri\width} {\trimzoomuppermri\height}},clip,cfbox=red 0.5pt 0pt]{f_tilde0_Nlines=16_idimg=20_fixed.png}}
            };
            \draw [stealth-,red] (-0.4,0.35) -- (0.27,-0.25);
        \end{tikzpicture}
        \\ $15.17$dB} &
        \makecell{
        \begin{tikzpicture}
            \node[inner sep=0pt] (img) at (0,0) {
                \includegraphics[width=\subfigwidth]{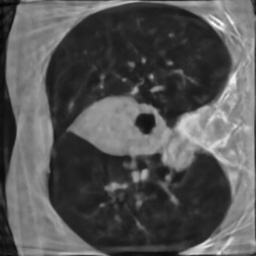}%
                \llap{\adjincludegraphics[width=0.083\textwidth,trim={{\trimzoomleftct\width} {\trimzoomlowerct\height} {\trimzoomrightct\width} {\trimzoomupperct\height}},clip,cfbox=red 0.5pt 0pt]{fig1c_d_25.70.png}}
            };
            \draw [stealth-,red] (-0.35,0.1) -- (0.27,-0.25);
        \end{tikzpicture}
        \\ $25.70$dB} &
        \makecell{
        \begin{tikzpicture}
            \node[inner sep=0pt] (img) at (0,0) {
                \includegraphics[width=\subfigwidthblur]{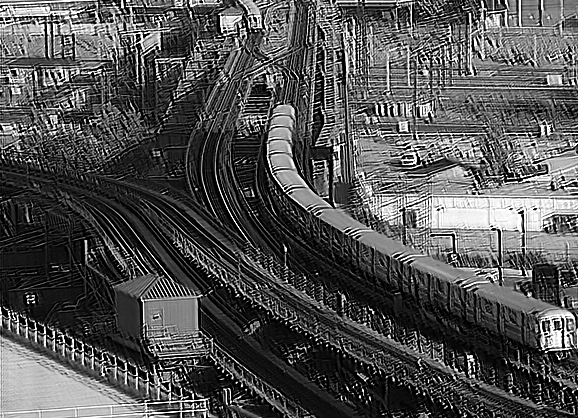}%
                \llap{\adjincludegraphics[width=0.083\textwidth,trim={{\trimzoomleftblur\width} {\trimzoomlowerblur\height} {\trimzoomrightblur\width} {\trimzoomupperblur\height}},clip,cfbox=red 0.5pt 0pt]{f_tilde0_coco_idimg=10_seed=7_noiselevel=0.0001_fixed.png}}
            };
            \draw [stealth-,red] (-0.3,1.3) -- (0.85,-0.15);
        \end{tikzpicture}
        \\ $14.54$dB} \\

        \midrule
        Test case &
        MRI &
        CT &
        Deblurring \\
    \end{tabular}
    \caption{Examples of the issues addressed in this paper. \emph{1st row:} description of the forward operators parameterized by $\thetaz_0$ and $\thetaz_1$. \emph{2nd row:} pseudo-inverse reconstruction of $\yz_0=\Pc(\Az(\thetaz_0)\xz)$ for MRI and CT and the blurry image $y_0$ for deblurring. \emph{3rd row:} reconstruction with no model or training mismatch. \emph{4th row:} reconstruction with a training mismatch. \emph{Last row:} reconstruction with a model mismatch (blind). All the models are an unrolled ADMM trained on $\Az(\thetaz_0)$. The reconstruction PSNR is provided below each image.}
    \label{tab:exampleissues}
\end{figure}

While model-based reconstruction networks provide state-of-the art results in a large panel of applications, it is now well established that they suffer from a \emph{lack of adaptivity}.
This means that  a network trained for a specific operator $\Az(\thetaz_0)$ may have a significant performance drop if used for another operator $\Az(\thetaz_1)$. This drop can be evaluated as follows.
Let $\thetaz_0\neq \thetaz_1$ denote two different operator parametrizations. Let $\yz_0=\mathcal{P}(\Az(\thetaz_0)\xz)$ and $\yz_1=\mathcal{P}(\Az(\thetaz_1)\xz)$. Assume that $\wz_0^\star$ and $\wz_1^\star$ are the weights of a reconstruction network optimized for $\Az(\thetaz_0)$ and $\Az(\thetaz_1)$ respectively. We compare the quality of $\Nc(\wz_0^\star,\Az(\thetaz_0), \yz_0)$ and $\Nc(\wz_1^\star,\Az(\thetaz_0), \yz_0)$ in the third and fourth rows of Fig.~\ref{tab:exampleissues}. Observe the significant performance difference.

To avoid this pitfall, we propose to train the network by minimizing~\eqref{eq:main_idea_of_the paper} instead of~\eqref{eq:standard_training_procedure}. \rev{After providing some theoretical hints on why a favorable behavior may occur,} we will carefully evaluate the performance of the resulting networks in Section~\ref{sec:results} for MR image reconstruction from under-sampled data, \rev{CT imaging} and image deblurring.
We conclude that this learning approach yields a reconstruction network which is significantly more stable to variations of the forward operator.
In addition, the performance of an unrolled network trained on a restricted family is only marginally worse than that of a network that would be trained and used for a single operator. 
It therefore provides a satisfactory answer to the adaptivity issue.
We also address several questions raised by our methodology. Can the unrolled network trained on a family extrapolate to unseen operators? How to sample the space of admissible operators? What is the gain of \rev{this} approach in comparison to more ``universal approaches'' such as plug-and-play (P\&P) priors?

\paragraph{\rev{Model mismatch issue}}

Assume that we observe $\yz_1=\mathcal{P}(\Az(\thetaz_1)\xz)$. Unfortunately, we only have access to an approximate knowledge $\Az(\thetaz_0)$ of the \rev{true} forward model \rev{$\Az(\thetaz_1)$}.
This can be due to an imprecise calibration of the sensing device or to the motion of a patient in a scanner for instance. 
\rev{We then face a blind inverse problem.}
A problem solved with such a model mismatch (i.e. with the operator $\Az(\thetaz_0)$ in place of $\Az(\thetaz_1)$), can lead to catastrophic reconstruction results, as illustrated in the last row of Fig.~\ref{tab:exampleissues}.

The second contribution of this work is to propose a systematic approach \rev{called ``deep unrolled prior''} to recover an estimate $\hat\thetaz_1$ of $\thetaz_1$ from the observation $\yz_1$. 
We show that unrolled networks trained on a family of forward models provide a powerful tool to solve several blind inverse problems. The idea is simply to minimize the data consistency error
\begin{equation}\label{eq:blindinvmintheta}
    \hat\thetaz\in\argmin_{\thetaz \in \Theta} \frac{1}{2}\Vert \Az(\thetaz) \Nc\left(\wz, \Az(\thetaz), \yz\right) - \yz \Vert_2^2.
\end{equation}
The reconstructed image $\hat \xz \eqdef \Nc\left(\wz, \Az(\hat \thetaz), \yz\right)$ is defined as the output of the unrolled neural network.
This consistency principle is spread massively in the literature of blind inverse problems \rev{and usually appears when constructing maximum a posteriori estimates}. The main contribution here is to plug it with a specific training procedure on a family of forward operators.

\section{Related works}

\paragraph{Regularization theory}

From a historical perspective, the first inverse problem solvers were based on simple inverses or approximate inverses of $\Az(\thetaz)$.
This approach provides low quality results when the matrix $\Az(\thetaz)$ has a non trivial kernel or when the conditioning number of $\Az^*(\thetaz)$ is high.
In those cases, it is critical to use regularization terms.
For long ($\sim$ 1960-2000), simple quadratic terms (Tikhonov) dominated the scientific landscape.
Around 1990, a second research trend appeared with convex, nonlinear regularizers such as total variation \cite{rudin1992nonlinear}. This area culminated with the development of the compressed sensing theory \cite{candes2006robust,lustig2005faster}.

\paragraph{Learned reconstruction}

\rev{In the 2010's learned regularizers such as variational networks emerged \cite{hammernik2018learning,hertrich2022wasserstein,goujon2022neural,altekruger2022patchnr}. They can be convex or nonconvex and can come with nice theoretical guarantees (e.g. robustness and stability) developed in the frame of compressed sensing. They apply seamlessly to a large variety of inverse problems.}

Starting from 2015, impressive performance gains have occurred with the advent of neural networks. They seem able to replace the initial methods in a growing number of technologies \cite{wang2018image}.
There are two main approaches to attack reconstruction problems using machine learning \cite{arridgesolving}. 
A first solution is \emph{end-to-end networks} where the neural network is agnostic to the operator $\Az(\thetaz)$. It gets trained through pairs $(\yz_i, \xz_i)$ generated with the model \eqref{eq:forward_model}. A popular example is AUTOMAP \cite{zhu2018image}. In this algorithm, the network needs to infer the forward model from the training data. This usually requires a huge amount of training data for large $M$ and $N$.

The other possibility is \emph{model-based} reconstruction networks that are defined as mappings of the form \eqref{eq:reconmapping}.
They are often praised for the fact that they require less training data and benefit from a higher interpretability.
Two popular approaches among this class are:
\begin{itemize}
  \item \emph{Denoising nets}: The reconstruction network performs a rough inversion followed by a denoising network such as a U-Net, to remove the remaining artifacts, see e.g. \cite{jin2017deep}.
  \item \emph{Unrolled nets}: Many efficient iterative methods have been developed to solve convex optimization problems (proximal gradient descent, Douglas-Rachford, ADMM, Primal-Dual, ...) \cite{combettes2011proximal}. They have the general form:
  \begin{equation}
  \xz_{k+1} = \prox_{R} \left( M(\Az(\thetaz), \yz, \xz_k) \right), \label{eq:iterative_scheme}
  \end{equation}
  for $k=1$ to $K\in \N$. The mapping $M$ can be interpreted as a crude way to invert the operator, in the sense that $\Az(\thetaz)M(\Az(\thetaz), \yz, \xz_k) \simeq \yz$. The term $\prox_{R}$ can be interpreted as a way to regularize (denoise) the remaining artifacts. The so-called P\&P priors \cite{venkatakrishnan2013plug} fit in this category.

  The unrolled networks draw their inspiration from \eqref{eq:iterative_scheme}. They consist in replacing the handcrafted or learned proximal operator $\prox_{R}$ by a sequence of neural networks \rev{$(\Dc(\wz_k,\cdot))_{1\leq k \leq K}$} promoting an output $\xz_K$ similar to the training images. 
  The difference with the P\&P priors is that the weights $\wz_k$ are trained specifically for a given operator $\Az(\thetaz)$.
  Examples of approaches in this category include \cite{sun2016deep,diamond2017unrolled,adler2017solving,zhang2018ista,dong2018denoising,adler2018learned,aggarwal2018modl,hammernik2019sigma,li2019algorithm}. These algorithms are currently among the most efficient for MRI reconstruction \cite{muckley2021results}.
\end{itemize}
For completeness, let us mention that a popular alternative consists in synthesizing the images $\xz$ with generative models \cite{bora2017compressed,asim2020blind}. Compared to the approaches mentioned above, it typically suffers from a higher computational cost. Indeed, a gradient descent in the latent space needs to be performed. In addition, specific care must be taken to handle images living outside the range of the generator. Hence, we will not consider this approach further in this work.

\paragraph{Adaptivity}

Neural network reconstructions can suffer from severe instabilities. 
This issue was notably discussed in \cite{antun2020instabilities}. 
The authors show that well chosen additive noise (an adversarial attack) or modifications of the forward operator can lead to disastrous hallucinations for some specific architectures. 
This problem was also studied experimentally in \cite{genzel2022solving}. There, the authors have shown that careful training procedures could fix many issues and yield robust and state-of-the-art reconstruction results, with a stability on par with handcrafted methods. 
\rev{Yet, it should be noted \cite{gottschling2020troublesome} that there is a fundamental bottleneck in the resolution of severely ill-posed inverse problems. In fact, any attempt to solve some of their instances stably and accurately is doomed, since multiple plausible signals may live simultaneously in the kernel of the forward operator}.

A paper closely related to our work is \cite{gilton2021model}. The authors study the same robustness issue to model mismatches.
\rev{They} propose two distinct algorithmic approaches to address it.
The first one is called parameterize \& perturb by the authors. 
It suffers from an important drawback, which is the need to re-optimize the network weights for every new operator. 
It can therefore be slow at run time and we do not compare it in this paper.
The other approach is called Reuse \& Regularize (\emph{R\&R}). It consists in training a network for a given operator $\Az(\thetaz_0)$, and then use this network as a regularizer for another operator $\Az(\thetaz_1)$. This is done in an iterative procedure, accounting for the data consistency term $\|\Az(\thetaz_1) \xz - \yz\|_2^2$. The approach we propose in this paper is significantly lighter at run time: we just train the network once with a family of operators \rev{and use it for multiple operators.}

An older and popular alternative consists in replacing the proximal operator in \eqref{eq:iterative_scheme} by a denoiser. 
This approach is often called a plug-and-play (P\&P) prior \cite{venkatakrishnan2013plug}.
It was first used with hand-crafted priors \cite{gu2014weighted} and a significant performance boost occurred with the use of pre-trained neural networks among which we can cite \cite{ryu2019plug,zhang2021plug}.
\rev{In addition, let us mention that \cite{zhang2021plug} trains the denoiser with various noise levels (instead of forward operators). This makes it possible to fine-tune the regularization in the P\&P method.}
This approach has the huge asset of adapting painlessly to arbitrary inverse problems. 
We propose some comparisons and discuss the pros and cons of each approach in Section \ref{sec:results}.

\rev{Finally, let us point out that the idea of training solvers on families of operators is probably implemented already on a variety of neural networks. For instance, the SFTMD network in \cite{gu2019blind} is trained on a family as well.
Our main contribution here is a systematic empirical study of this methodology. }

\paragraph{Blind inverse problems}

Blind inverse problems are spread massively in applications.
The review papers \cite{kundur1996blind,campisi2017blind} provide a good idea of the wealth of results for the sole field of blind deconvolution and super-resolution.

A possibility is to design a two-step method. First an estimate of the forward operator is built. 
Second, this estimate is used in conjunction with the methods from the previous section. 
In some cases, it is possible to exploit some redundancy in the data to estimate the operator parameters. This is the case in parallel MRI, where the coil sensitivity maps can be estimated using only the low frequencies \cite{sodickson1997simultaneous,pruessmann1999sense,griswold2002generalized}.
When no redundancy is available, estimating the operator can be achieved by minimizing the discrepancy between the statistics of the acquired measurement and the statistics of the measurements generated by applying an operator to a ``natural'' signal. A good example in blind deblurring is the Goldstein-Fattal approach \cite{goldstein2012blur}, which analyzes the power spectrum of the blurry image. 
A few authors proposed to build an identification network that learns to identify the blur kernel \cite{schuler2015learning} or a blur parametrization \cite{sun2015learning,yan2016blind,chakrabarti2016neural,debarnot2022deep} from the blurry-noisy image. While this approach is cheap computationally, it requires an application specific design.

One of the most popular alternatives consists in minimizing a combination of a data fidelity term and a regularizing prior. This can be addressed through an alternate minimization between the image and the operator parametrization. Most of the literature suggests the use of hand-crafted priors on the unknown operator or on the image to recover (see e.g. \cite{chan1998total,fergus2006removing,krishnan2009fast,krishnan2011blind,xu2013unnatural,ahmed2013blind,pan2014deblurring,michaeli2014blind,pan2016blind,ren2016image,bai2018graph,Ljubenovi2019,chen2019blind,9878959} for blind deblurring, or \cite{riis2021computed,wang2022perfect,meng2023numerical} in CT imaging). 

While these approaches can provide excellent results, they are likely to be outperformed by neural network based approaches in a near future. Indeed, impressive performance has already been reached recently thanks to neural network based regularizers. Different strategies have been suggested, going from untrained networks (see \cite{Bostan:20} for an application in optics), generative models (see \cite{asim2020blind} for an application in blind deblurring), or unrolled networks (see \cite{lecouat_mairal,DBLP:conf/iccv/LecouatPM21} for an application to super-resolution from an image sequence).

The method advocated in our paper is close in spirit to the works in this latest category. It differs in the way the training is performed. \rev{In our work}, we first train an unrolled network on a family of forward operators, which allows fixing the weights once for all. We then minimize \eqref{eq:blindinvmintheta} in the space of parameters of the forward model. This methodology has various advantages:
\begin{itemize}
    \item Compared to untrained networks \cite{Bostan:20}, the method does not optimize the network weights to solve the problem, which is typically quite computationally heavy. It is therefore faster at evaluation time. In addition, it is adapted to a clearly defined image dataset.
    \item Methods based on generative models \cite{bora2017compressed,asim2020blind} may suffer from a significant drawback: the produced images necessary live in the range of the generator. To avoid this issue, a possibility is to add hand-crafted regularization terms such as total variation that allow extending the span of possible images \cite{asim2020blind}.
    \item In \cite{lecouat_mairal,DBLP:conf/iccv/LecouatPM21}, the neural network weights are trained directly to solve the blind inverse problem. This significantly limit the number of weights and iterations within the iterative procedure. In this paper, we propose to train the network beforehand, allowing to use arbitrary solvers and as many iterations as desired to find the parameter $\thetaz$.
\end{itemize}

\section{Preliminaries}

In this paper, we consider forward models of the form
\begin{equation}\label{eq:linear_forward_model}
    \yz = \Az(\thetaz)\xz + \bz
\end{equation}
where $\Az(\thetaz) \in \K^{M\times N}$ is a linear mapping either real ($\K=\R$) or complex ($\K = \C$). 
In all our experiments, we define $\bz$ as additive white Gaussian noise (complex for MRI) $\bz \sim \Nc(0,\sigma^2 \Id)$. 
The dependency of $\Az$ with respect to its parameter $\thetaz$ can be linear or nonlinear.
We let $N\in \N$ denote the number of pixels of the image $\xz$ with $N=N_x\times N_y$ for 2D images and $M$ is the number of measurements.

\subsection{Forward models}

\rev{
To illustrate our problem, we consider three important biomedical applications: parallel magnetic resonance imaging, computerized tomography and microscopy/astronomy. We provide a quick overview of the models below and a more precise mathematical description is given in Appendix \ref{app:precise_description}.}

\begin{figure}
    \centering
    \includegraphics[width=0.13\linewidth]{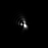}
    \includegraphics[width=0.13\linewidth]{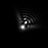}
    \includegraphics[width=0.13\linewidth]{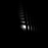}
    \includegraphics[width=0.13\linewidth]{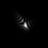}
    \includegraphics[width=0.13\linewidth]{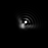}
    \includegraphics[width=0.13\linewidth]{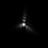}
    \includegraphics[width=0.13\linewidth]{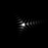}

    \includegraphics[width=0.13\linewidth]{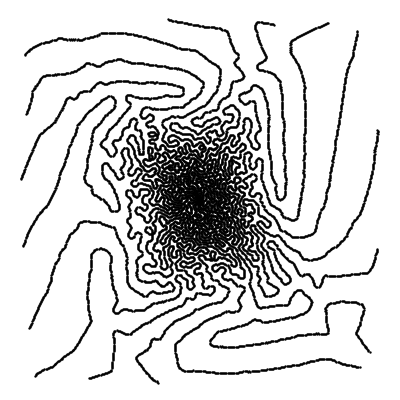}
    \includegraphics[width=0.13\linewidth]{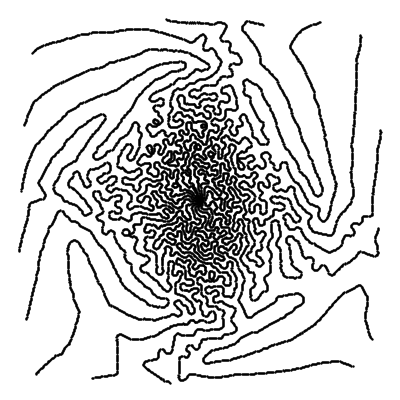}
    \includegraphics[width=0.13\linewidth]{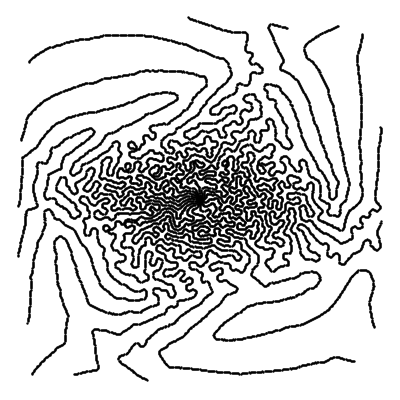}
    \includegraphics[width=0.13\linewidth]{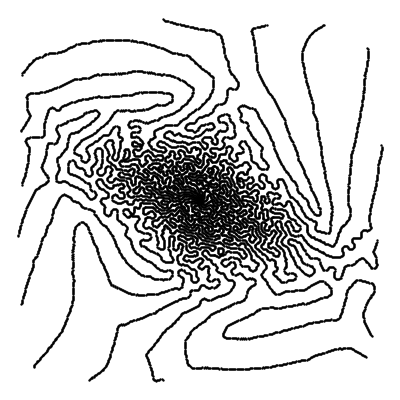}
    \includegraphics[width=0.13\linewidth]{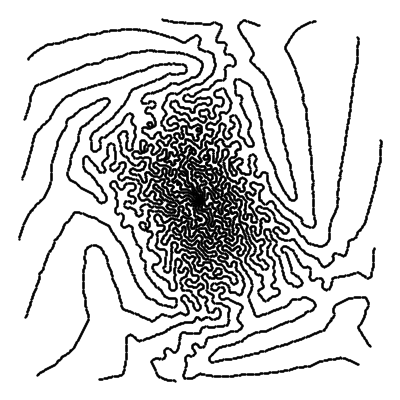}
    \includegraphics[width=0.13\linewidth]{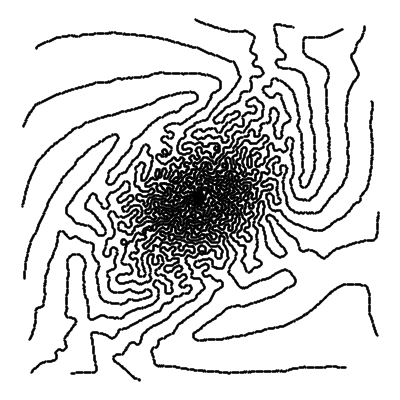}
    \includegraphics[width=0.13\linewidth]{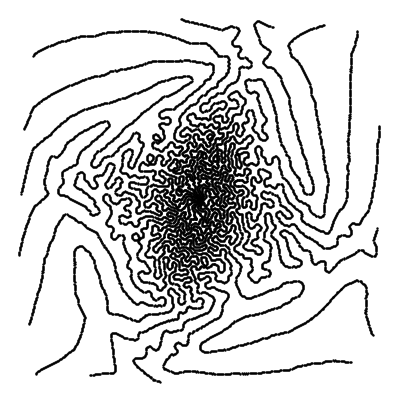}

    \caption{\rev{Top: examples of point spread functions generated with Fresnel diffraction theory. The pupil function is defined through a linear combination of 7 Zernike functions with random coefficients. The dependency on the coefficients is highly nonlinear. All PSFs are realistic (e.g. non-negative and bandlimited).
    Bottom: examples of sampling schemes in parallel MRI used in this work. All sampling schemes are realistic and can be implemented on an actual scanner. They include realistic physical constraints of speed and acceleration (maximum gradient amplitude and slew rate). \label{fig:random_psf}}}
\end{figure}

\rev{\paragraph{Parallel Magnetic Resonance Imaging} In this application, $\Az(\thetaz)$ is the product of a partial non uniform Fourier transform with a set of diagonal matrices encoding the ``sensitivities'' of reception coils around the object to image. The samples in the Fourier domain, also denoted k-space, are located along a smooth trajectory.
The reconstruction network should adapt to:
\begin{itemize}
     \item different sampling trajectories,
     \item different sensitivity maps (which are smoothly varying multipliers),
     \item the effect of imperfect gradient coils/ off-resonance effects that deteriorate the trajectory.
 \end{itemize} 
In our experiments, we consider a subsampling ratio of $4$ and $10$, meaning that $M=N/4$ or $M=N/10$ respectively.
The parameter $\thetaz$ encodes all the parameters above. 
The sensitivity maps and the trajectory perturbations are usually unknown, making MRI reconstruction a blind inverse problem. Examples of realistic sampling trajectories used in this work are displayed in Fig. \ref{fig:random_psf}, bottom.}

\rev{\paragraph{Computerized tomography} We consider parallel beam X-ray computerized tomography. 
It consists in probing line integrals of an object along a set of parallel lines that may be rotated and shifted. In this application the parameter $\thetaz$ represents the angles and shift at origin of the lines. The problem becomes blind if the object to image moves during the scan.}

\rev{\paragraph{Deblurring in optics} The most common way to parametrize the Point Spread Function (PSF) of an optical system in optics is by using Fresnel diffraction theory \cite{born2013principles}. In this theory, the PSF is entirely determined by the pupil function, which is a complex function defined over the objective aperture. For a circular aperture, the pupil function can be expanded with Zernike polynomials, which are orthogonal polynomials over the disk \cite{noll1976zernike,lakshminarayanan2011zernike}. The parameter $\thetaz$ coincides with the coefficients of this linear decomposition. The problem is blind whenever the PSF is unknown.  In our experiments, we consider a linear combination of $7$ Zernike polynomials. Examples of random PSFs generated through this model are displayed in Fig. \ref{fig:random_psf}, top.}

\subsection{Model-based reconstruction networks\label{sec:archictectures}}

In this paragraph, we detail the neural network architectures considered in this work for the numerical experiments.
\rev{In all the following, $\Dc:\R^D\times \K^N \to\K^N$ denotes a neural network with weights $\wz\in \R^D$ and input signal $x\in \R^N$.}
The letter $\Dc$ stands for denoising, since the goal of this network is to remove artifacts on $\xz$ remaining after inversion of the forward model.

\subsubsection{Inversion + denoising network}
Possibly the simplest way to construct an operator-aware reconstruction network is to consider a mapping $\Nc^d$ of the form:
\begin{equation} \label{def:denoising_network}
\Nc^d(\wz,\Az(\thetaz),\yz) \eqdef \Dc\left(\wz,\Az(\thetaz)^\dagger\yz\right),
\end{equation}
where $\Az(\thetaz)^\dagger$ is the pseudo-inverse of $\Az(\thetaz)$.
The idea is simply to roughly invert the model and train a single denoising network $\Dc$ to remove the artifacts \cite{jin2017deep}.
This type of network was one of the earliest ones. 
We will consider this architecture only for a single MRI experiment due to its overall poor performance.

\subsubsection{Unrolled ADMM} 

The unrolled ADMM network is an efficient architecture providing results close to the state-of-the-art in a number of applications. 
It takes the form (see e.g. \cite{ng2010solving}):
\begin{align*}
    \xz_0 &= 0 \quad \mbox{and} \quad  \muz_0 = 0 \\ 
    \zz_{k+1} &= \left[ \Az^*\Az+\rev{\beta_k}\Id \right]^{-1}\left( \Az^*\yz+\rev{\beta_k} \xz_k-\muz_k \right) \\
    \xz_{k+1} &= \Dc\left(\wz_k, \zz_{k+1}+\frac{\muz_k}{\rev{\beta_k}} \right) \\
    \muz_{k+1} &= \muz_k+\rev{\beta_k}\left( \zz_{k+1}-\xz_{k+1} \right).
\end{align*}
This sequence runs for $K$ iterations and the result is denoted $\Nc^a: (\wz, \Az, \yz) \mapsto \xz_K$ \rev{with $\xz_K$ the final iterate and ``a'' in $\Nc^a$ stands for ADMM}.
\rev{The number $K$ will be set equal to $5$ when the weights $\wz$ need to be trained. 
This is mostly due to memory and computing time limitations.
This ADMM based architecture can also be used as a P\&P algorithm, in which case, a higher number of iterations can be considered to obtain the best possible signal-to-noise ratio.} The parameter $\beta_k$ is a penalty parameter, \rev{which can vary from one iteration to the next. We use the update rule proposed by \cite{zhang2021plug} for the P\&P algorithms in all experiments}. 
The weights $\wz$ to be trained are $\wz = [\wz_0,\hdots, \wz_{K-1}]$. They differ at each iteration for the unrolled networks and are identical in the P\&P networks.

\rev{\subsubsection{Denoising network architecture}}

\rev{All our experiments are achieved with a \emph{fixed denoising architecture} $\Dc$. 
We choose the so-called DRUNet network \cite{zhang2021plug} (for Denoising Residual U-Network).
This network is the current state-of-the-art when used within P\&P algorithms. One of its important assets is its ability to accommodate for different noise levels. The idea is to set one of the input channels as a constant image with a value equal to the standard deviation of the noise.
This is an important feature for P\&P algorithms, which depend on a parameter describing the noise level. 
The same advantage applies to unrolled networks.
The noise level is a user-defined parameter that can be changed to vary the regularization level depending on the application.
We decided to use a single denoising architecture in our experiments for the following reasons:
\begin{itemize}
    \item The network is currently the state-of-the-art for the field of P\&P methods. It therefore makes the comparison with the P\&P methods more relevant.
    \item Compared to other architectures we have tried, the training stage was easier and the performance higher.
    \item We want to simplify the message by avoiding too many experiments and by comparing the different methods using only a single architecture.
    \item One single training of each network is already about a week of computation on an A100 Nvidia graphics card and we want to reduce the overall computing time for this paper.
  \end{itemize}
}

\section{Training with an operator distribution}\label{sec:trainingfamily}

\rev{
In most existing approaches, networks are trained by minimizing the empirical risk for a given forward model $\Az(\thetaz_0)$ as in  \eqref{eq:standard_training_procedure}. 
Instead, we propose to minimize the risk  over an operator distribution as in \eqref{eq:main_idea_of_the paper}. 
In this section, we explain a few differences between both approaches.

To begin with, notice that the pre-image by $\Az(\thetaz)$ of the measurement vector $\yz=\Az(\thetaz)\xz+\bz$ is given by
\begin{equation}\label{eq:pseudo_inverse_terms}
    \Az(\thetaz)^{-1}\yz = \left\{\xz + \kz + n, \kz\in \mathrm{ker}(\Az(\thetaz)), n = \Az(\thetaz)^\dagger\bz \in \rev{\mathrm{ker}(\Az(\thetaz))^\perp}\right\}.
\end{equation}
The vector $\wz$ is a correlated noise with a correlation that depends on $\Az(\thetaz)$.
Hence, the reconstruction network $\Nc$ should serve two purposes: 
\begin{enumerate}
    \item Recover the missing data $\kz$ in the kernel of $\Az(\thetaz)$.
    \item Remove the correlated noise $n$.
\end{enumerate}
Each of these two tasks is clearly highly dependent on $\Az(\thetaz)$.
We explain below that training a network on a single operator may result in some overfitting for the operator $\Az(\thetaz_0)$ and to a lack of generalization to other operators. This problem is strongly mitigated by training the network on a family. 
}

\rev{
\subsection{Large operator families are better in an ideal world \label{sec:ideal_world}}

We take a Bayesian viewpoint and assume that $\xb$ is a random vector in $\K^{N}$ with probability distribution measure $\mu_x$. 
We also see $\Ab$ as a random operator in $\K^{M\times N}$ with distribution $\mu_{A}$. 
Finally, we consider the random measurement vector $\yb$ generated through the forward model \ref{eq:forward_model}. 
Alternatively, we can see the parameter $\thetab\in \R^P$ as a random vector with probability distribution measure $\mu_{\theta}$ and we construct the random operator $A(\thetab)$, i.e. pushforward $\thetab$ through the mapping $\Az(\cdot)$.
For instance, the traditional training procedure \eqref{eq:standard_training_procedure} for a given operator $\Az(\thetaz_0)$ consists in assuming that $\mu_{A}=\delta_{\Az(\thetaz_0)}$ or equivalently $\mu_{\theta}=\delta_{\thetaz_0}$.

\paragraph{MMSE estimator}
In this framework, we may want to construct the Minimum Mean Square Error (MMSE) estimator.
The Mean Square Error (MSE) of an estimator $\hat \xz:\K^{M\times N} \times \K^M$ is a measure of performance defined by:
\begin{equation}\label{def:MSE}
    \MSE(\hat x) \eqdef \E_{\xb, \Ab, \yb}\left[ \|\hat \xz(\Ab,\yb) - \xb\|_2^2\right].
\end{equation}
The conditional MSE is defined by:
\begin{equation}\label{def:condMSE}
    \MSE(\hat x | A,y) \eqdef \E\left[ \|\hat \xz(\Ab,\yb) - \xb\|_2^2 | \Ab = A, \yb = y \right].
\end{equation}
Any estimator that achieves the minimum MSE is called an MMSE estimator.
It is defined for (almost) all pairs $(\Az, \yz)\in \K^{M\times N}\times\K^M$ by:
\begin{equation*}
    \hat \xz_{\MMSE}(\Az,\yz) \eqdef \argmin_{\hat \xz \in \K^N}\MSE(\hat x | A,y) .
\end{equation*}
An important property of this estimator is that it can be expressed as the following conditional expectation \cite{kay1993fundamentals}:
\begin{equation*}
  \hat \xz_{\MMSE}(\Az, \yz) = \E\left[ \xb | \Ab = \Az, \yb=\yz\right].
\end{equation*}
By construction, this estimator is the best we can hope for, in average for a given distribution of triplet $(\xb, \Ab, \yb)$.

\paragraph{Perfectly trained neural networks are MMSE estimators}
Notice that the risk $E$ defined in \eqref{eq:main_idea_of_the paper} coincides with the MSE:
\begin{equation}\label{eq:riskEisMSE}
    E(\wz) = \MSE(\Nc(\wz, \cdot, \cdot)).
\end{equation}
Therefore, training a neural network amounts to finding the MMSE estimator among the family of estimators
\begin{equation*}
    \Fc\eqdef\left\{ \Nc(\wz, \cdot, \cdot), \wz\in \R^D\right\}.
\end{equation*}
To better understand the difference between training a network on a single operator or on a distribution we may make the following simplifying assumption.
\begin{assumption}[Zero approximation and optimization errors\label{ass:ideal}]
    We assume that:
    \begin{itemize}
         \item the family $\Fc$ contains the MMSE estimator $\hat \xz_{\MMSE}$ for the distributions $\mu_{\theta_0}$ and $\mu_{\Ac}$.
         \item the optimizer returns a global minimizer of the risk in \eqref{eq:main_idea_of_the paper}.
    \end{itemize} 
\end{assumption} 
In the language of \cite{bottou2007tradeoffs}, this means that the approximation and optimization errors vanish. 
Obviously, this is not realistic in general, but many recent experiments show that it can be considered approximately correct for large overparameterized networks (see e.g.  \cite{belkin2021fit}).
Under those assumptions, the following straightforward result shows that it can only be beneficial to train a network on a distribution of operators with a large support.
\begin{proposition}
Let $w_0$ denote the weights optimized using a single operator $A(\theta_0)$.
Let $w_{\mu_A}$ denote the weights optimized using the distribution of operators $\mu_A$.
Let $\Ac \eqdef \mathrm{supp}(\mu_A)$ denote the family of operators that was used for training. 
Under Assumption \ref{ass:ideal}, we get:
\begin{align*}
\Nc(\wz_{\mu_A},\Az, \yz) &= \hat \xz_{\MMSE}(\Az, \yz) \quad \mbox{for almost all} \quad \Az\in \Ac, \yz \in \K^M. \\
\Nc(w_0,\Az(\thetaz_0), \yz) &= \hat \xz_{\MMSE}(\Az(\thetaz_0), \yz) \quad \mbox{for almost all} \quad \yz\in \K^M.
\end{align*}
However, $\Nc(w_0,\Az,\yz)$ may differ from $\hat \xz_{\MMSE}(\Az,\yz)$ whenever $\Az\neq \Az(\thetaz_0)$.
\end{proposition}
The above proposition is straightforward. It is a simple consequence of the fact that the weights are optimized to minimize the MSE. 
It tells us that the neural network $\Nc(\wz_{\Ac},\cdot,\cdot)$ trained on an operator distribution coincides with the MMSE for almost every operator on the support. Hence, under Assumption \ref{ass:ideal}, it is as good as can be for every operator seen during training.
In particular it implies that $\Nc(\wz_{\Ac},\Az(\thetaz_0),\yz) = \Nc(w_{0},\Az(\thetaz_0),\yz)$ if $\Az(\thetaz_0) \in \Ac$.
This means that there is no disadvantage to train the network on a family, even for the specific operator $\Az(\thetaz_0)$. This phenomenon will be (nearly) confirmed later in the numerical experiments.

On the other hand, nothing can be said for a network trained on a single operator $\Az(\thetaz_0)$ when applied to another operator $\Az\neq \Az(\thetaz_0)$. There, we need to rely on the generalization capacity of the network. This capacity looks really arbitrary since a single operator was seen during training. 
This is the most likely explanation for the lack of adaptivity that was observed in Fig. \ref{tab:exampleissues}.
To sum up, \emph{under the idealist hypothesis \ref{ass:ideal}, it can only be beneficial to train the network on the largest possible family of operators.}

Finally, let us mention that under Assumption \ref{ass:ideal}, we could learn the prior $\mu_x$ exactly.
This would make it possible to sample the posterior distribution $\mu_{x|(A,y)}$ directly \cite{laumont2022bayesian,song2023pseudoinverseguided,chung2023diffusion}. Hence, the MMSE estimator could be accessed for \emph{every} operator $A\in \K^{M\times N}$ (using, e.g. Langevin dynamics).
However, this would come at the price of a significantly increased computational time to solve each inverse problem instance.  
}

\rev{
\subsection{The possible downsides}

In the previous section, we made the following unrealistic assumptions:
\begin{enumerate}
    \item[i)] The family $\Fc$ is so large that it contains the MMSE estimator. In practice we use structured convolutional neural network which cannot approximate arbitrary functions and should account for approximation errors. 
    \item[ii)] The expectation with respect to $\mu_x$ can be evaluated. In most applications, we can only resort to a finite size dataset and to the minimization of the empirical risk.
    \item[iii)] The optimization routine returns the global minimizer. In most cases, the stochastic gradient descents used for training only return approximate critical points. 
\end{enumerate}

Each of the above points makes the above analysis imprecise.
In particular, if we only assume points ii) and iii) to hold, we get:
\begin{equation*}
\MSE( \Nc(w_0 , \Ab, \yb) | A_0, y) = \inf_{w\in \R^D} \MSE( \Nc(w , \Ab, \yb) | A_0, y) \leq \MSE( \Nc(w_{\mu_A} , \Ab, \yb) | A_0,  y).
\end{equation*}
In general, the inequality above is strict since it is much easier to approximate the mapping $\hat x_{\MMSE}$ pointwise on the domain $\{(A_0,y), y\in \K^M\}$ than on the whole domain $\{(A,y), A\in \Ac, y\in \K^M\}$.
Hence, the lack of expressiveness of the network $\Nc$ will -- in general -- result in a performance decrease for the specific operator $A_0$. We will see empirically that this decay is moderate for the 3 applications considered in this paper. Providing bounds on this decay is an intricate issue that is left open for future works.
This simple observation however reveals a potential pitfall of the distribution training approach: the larger the family $\Ac$, the worst the performance for specific operators in $\Ac$. This shows that there is an adaptivity/performance trade-off, especially if the family $\Fc$ lacks expressiveness.

}

\subsection{Choosing distributions of operators}

\rev{
Choosing a proper family and distribution of operators obviously depends on each application. 
Ideally, this distribution should reflect the real distribution of the underlying imaging system. 
Unfortunately, this is often hard to characterize.
As mentioned in the previous section, the main feature to consider is the family of operators $\Ac$ seen during training, i.e. the support of the distribution $\mu_A$.
This family should be sufficiently large to reflect any operator that could arise in practice. 
Once this family is characterized, it is possible to sample it as uniformly as possible. 
Overall, the choice of an operator distribution is nontrivial and should rely on an expert knowledge of the imaging system. 
We detail how we addressed this question for the three applications below.}

\subsubsection{Magnetic Resonance Imaging}

In this modality, the family of forward operators is constructed by considering different sampling schemes and sensitivity maps.

\paragraph{Sampling schemes}
We propose to generate random sampling schemes  $\xib$ following the ideas from \cite{boyer2016generation,chauffert2017projection,lazarus2019sparkling}. 
The principle is to design a scheme that fits a target probability measure $\rho:\R^2\to \R_+$.  
To this end, we define
\begin{equation}\label{eq:sampling_pattern}
\xib(\rho) \eqdef \argmin_{\xib \in \Xi} \dist\left(\frac 1 M\sum_{m=1}^M \delta_{\xib_m},\rho\right),
\end{equation}
where $\dist$ is a discrepancy between probability measures and $\Xi\subseteq \R^{2\times M}$ is a set that describes the admissible trajectories from the scanner.

\begin{figure*}
    \centering
    \begin{subfigure}[b]{0.115\textwidth}
        \centering
        \includegraphics[width=\textwidth]{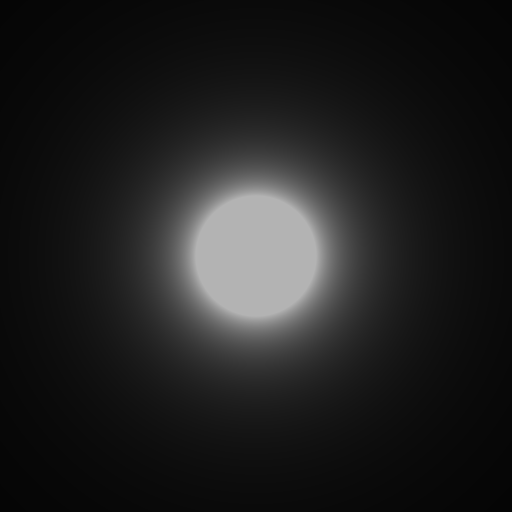}\vspace{0.05cm}\\
        \includegraphics[width=\textwidth,trim={10pt 10pt 10pt 10pt},clip]{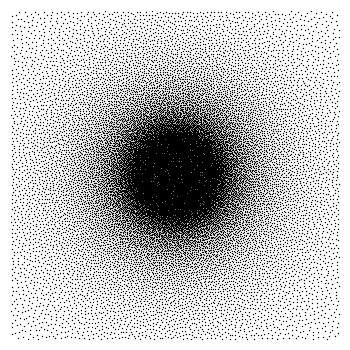}
        \caption{$\bigcirc$}\label{fig:ref}
    \end{subfigure}\ 
    \begin{subfigure}[b]{0.115\textwidth}
        \centering
        \includegraphics[width=\textwidth]{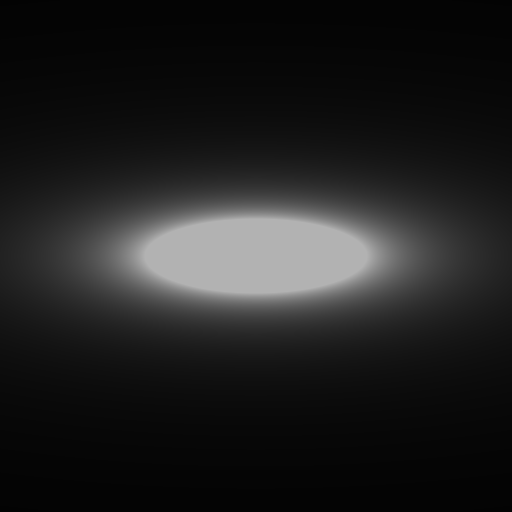}\vspace{0.05cm}\\
        \includegraphics[width=\textwidth,trim={10pt 10pt 10pt 10pt},clip]{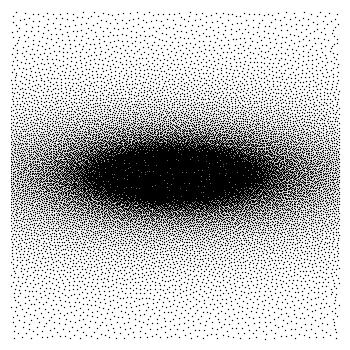}
        \caption{$-$}\label{fig:hor}
    \end{subfigure}\ 
    \begin{subfigure}[b]{0.115\textwidth}
        \centering
        \includegraphics[width=\textwidth]{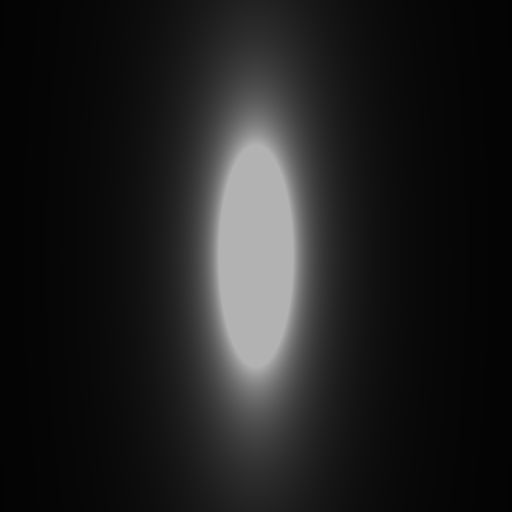}\vspace{0.05cm}\\
        \includegraphics[width=\textwidth,trim={10pt 10pt 10pt 10pt},clip]{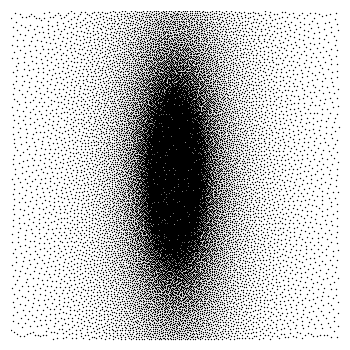}
        \caption{$|$}\label{fig:vert}
    \end{subfigure}\ 
    \begin{subfigure}[b]{0.115\textwidth}
        \centering
        \includegraphics[width=\textwidth]{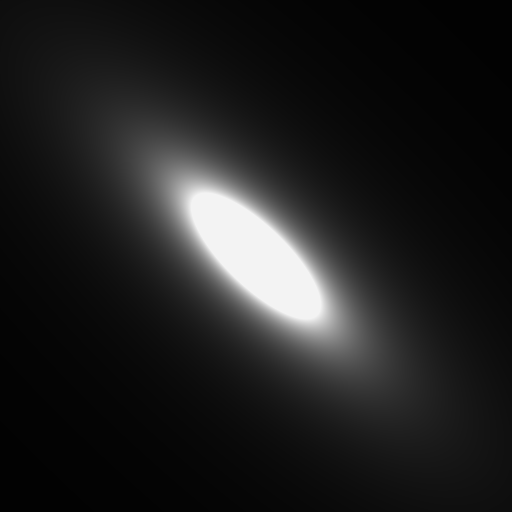}\vspace{0.05cm}\\
        \includegraphics[width=\textwidth,trim={10pt 10pt 10pt 10pt},clip]{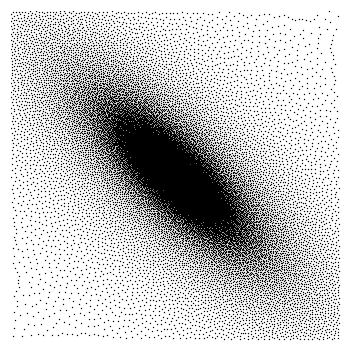}
        \caption{}\label{fig:sample0}
    \end{subfigure}\ 
    \begin{subfigure}[b]{0.115\textwidth}
        \centering
        \includegraphics[width=\textwidth]{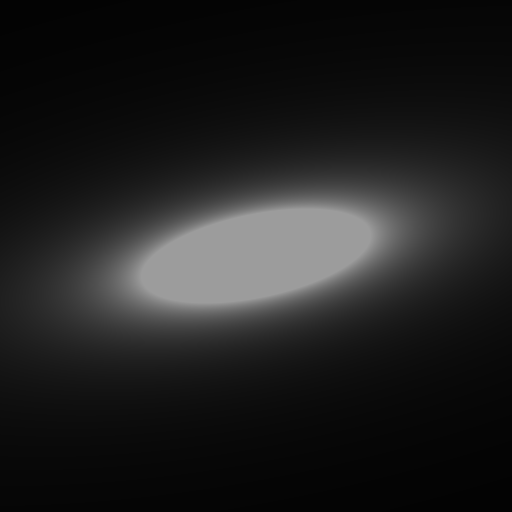}\vspace{0.05cm}\\
        \includegraphics[width=\textwidth,trim={10pt 10pt 10pt 10pt},clip]{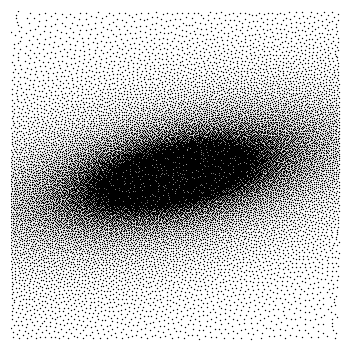}
        \caption{}\label{fig:sample2}
    \end{subfigure}\quad
    \begin{subfigure}[b]{0.115\textwidth}
        \centering
        \includegraphics[width=\textwidth]{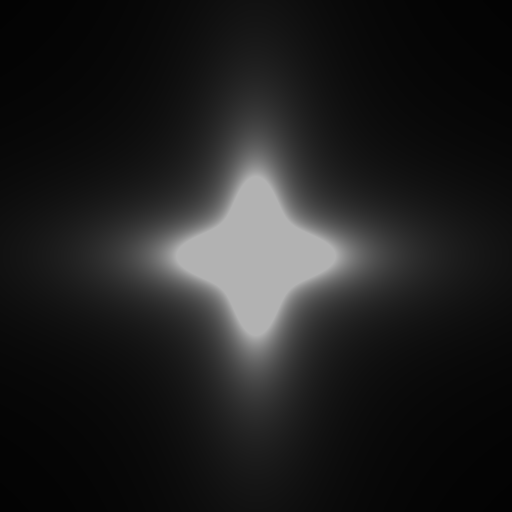}\vspace{0.05cm}\\
        \includegraphics[width=\textwidth,trim={10pt 10pt 10pt 10pt},clip]{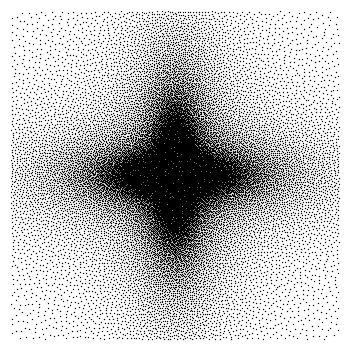}
        \caption{$+$}\label{fig:axes}
    \end{subfigure}\ 
    \begin{subfigure}[b]{0.115\textwidth}
        \centering
        \includegraphics[width=\textwidth]{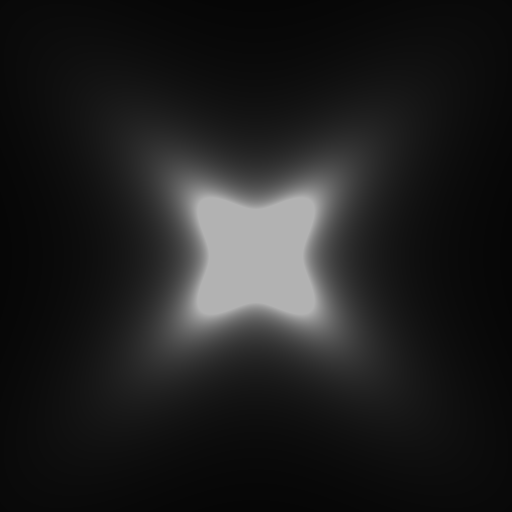}\vspace{0.05cm}\\
        \includegraphics[width=\textwidth,trim={10pt 10pt 10pt 10pt},clip]{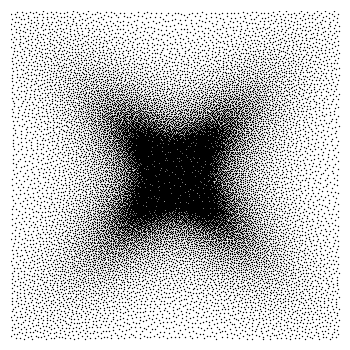}
        \caption{$\times$}\label{fig:diag}
    \end{subfigure}%
    \caption{Densities (top) and corresponding sampling schemes (bottom).
    Fig.~\ref{fig:ref}, \ref{fig:hor}, \ref{fig:vert}, \ref{fig:sample0} and \ref{fig:sample2} belong to the family $\Ac$. Fig.~\ref{fig:axes} and \ref{fig:diag} (crosses) do not. Notice that the sampling patterns are diverse with significant differences from one to the other.}
    \label{fig:scheme_and_density}
\end{figure*}

Following \cite{gossard2022bayesian}, we generate random target densities $\rho$ as anisotropic power decaying distributions.
They are parameterized by a random vector $\lambdab$ that encodes the density at origin, the anisotropy and the power decay law.
To avoid solving \eqref{eq:sampling_pattern} at training time, we have pre-computed $1000$ sampling patterns. The corresponding vectors $\lambdab$ have been generated by using a max-min sampling (see \cite{pronzato2017minimax,debarnot2022deep}) of a set of an admissible set of parameters $\Lambda$. We refer to \cite{gossard2022bayesian} for more details. Examples of densities and sampling patterns $\xib(\rho(\lambdab))$ without constraints are displayed in Fig.~\ref{fig:ref}-\ref{fig:sample2}. \rev{Notice that we did not include trajectory constraints for generating this figure. They are taken into account for the blind inverse problem part.}

\paragraph{Sensitivity maps}

As for the sensitivity maps, we used real estimates generated using the fastMRI database \cite{zbontar2018fastmri}. We first estimate them using a standard approach \cite{griswold2002generalized} and then project the estimates onto the span of a parametrization composed of thin plate splines.
At training time, they are associated to the corresponding training pairs.

\paragraph{Trajectories filtering}

We did not include the trajectory perturbation effect (convolution with $\hb(\omegab)$) at training time.

\subsubsection{Computerized tomography}

In this modality, we assume that the distribution of projection angles follows a uniform distribution centered on a vector of regularly spaced angles \rev{$\alpha_0 = (-\pi/2, -\pi/2 + \pi/J, \hdots , \pi/2)$} (see the red lines in first row of Fig.~\ref{tab:exampleissues}) and shift at origin $s_0=0$.
Hence, we have $\alpha=\alpha_0+\alphab_\delta$ with $\alphab_\delta\sim \mathcal{U}\left([-1.37^\circ,1.37^\circ]^J\right)$ and the random shifts are $\ssb \sim \mathcal{U}\left([-2,2]^J\right)$.
These perturbations may reflect movements of the patient inside the scanner during the scan.

\subsubsection{Deblurring}

In this application, we vary the blur kernel (the PSF) by changing only the $4$-th to the $10$-th Zernike polynomials.
We set $\theta_1=\theta_2=\theta_3=0$ in \eqref{eq:zernike_parametrization}.
\rev{In the Noll nomenclature, $\theta_1$ coincides with the piston, which does not change the PSF, $\theta_2$ and $\theta_3$ are tilts, which just shift the PSF.
We want to discard those coefficients to avoid the usual translation ambiguity in blind deconvolution.} We let the coefficients $\theta_4$ to $\theta_{10}$ follow a uniform distribution in $[-0.15, 0.15]$.

\section{\rev{Deep unrolled prior}}
\rev{
Assume that $y$ is generated according to the forward model $\yz = \Pc(\Az(\bar \thetaz)\bar x)$, \emph{with an unknown parameter} $\bar \theta$.
In that case, we need to estimate both $\bar x$ and $\bar \theta$, or alternatively the operator $\Az(\bar \thetaz)$.

\subsection{The proposed principle}

After training a network $\Nc$ on a family $\Ac$, we get a weight vector $w_{\mu_A}$.
For any pair $(A,y)\in \K^{M\times N}\times \K^M$, we are therefore able to build an estimate $\hat x(A,y) = \Nc\left(w_{\mu_A}, A, \yz\right)$ of $\bar x$.
We propose to estimate $\bar \theta$ by solving the optimization problem~\eqref{eq:blindinvmintheta}:
\begin{equation}
    \hat\thetaz\in\argmin_{\thetaz \in \Theta} \frac{1}{2}\Vert \Az(\thetaz) \Nc\left(\wz, \Az(\thetaz), \yz\right) - \yz \Vert_2^2.
\end{equation}

In this formulation, we wish the measurements $y$ to be consistent with the recovered signal, i.e. $A(\theta)\hat x(A(\theta),y) \approx y$. 
This approach could be called \emph{deep unrolled prior}, since we use an unrolled network as a prior to solve a blind inverse problem. 
Contrarily to the popular unsupervised method called \emph{deep image prior} \cite{ulyanov2018deep} though, our unrolled network is trained in a supervised way. It is then used without supervision to find the parameter $\theta$ only.
This approach can be related to a Bayesian approach.

\paragraph{Relationship with a MAP approach}

One of the most popular estimators for blind inverse problems is the Maximum A Posteriori (MAP).
To make a link with this approach, let us assume that the forward model reads:
\begin{equation}
     \yb = \Ab \xb + \bb,
 \end{equation} 
 where $\bb\sim \mathrm{Normal}(0,\sigma^2 \Id)$. The random operator $\Ab$ is drawn according to a distribution $\mu_A\propto \exp(-R_A)$, where $R_A:\K^{M\times N}\to \R\cup\{+\infty\}$ is a regularizer on the operator domain. Similarly, the random image $\xb$ is drawn according to a distribution $\mu_x\propto \exp(-R_x)$, where $R_x:\K^{N}\to \R\cup\{+\infty\}$ is a regularizer on the image domain.
We also assume independence of $\Ab$, $\xb$ and $\bb$.

In that case, it is tempting to solve the MAP problem:
 \begin{align}
     &\argmax_{\substack{A\in \K^{M\times N} \\ \ x\in \K^N}} \P\left(\Ab=A,\xb=x|\yb = y\right) \nonumber \\
     & = \argmax_{\substack{A\in \K^{M\times N} \\ \ x\in \K^N}} \P\left(\yb = y|\Ab=A,\xb=x\right)\cdot\exp(-R_x(x))\cdot \exp(-R_A(A)) \nonumber \\
     &= \argmin_{\substack{A\in \K^{M\times N} \\ \ x\in \K^N}} \frac{1}{2\sigma^2} \|Ax-y\|_2^2 + R_A(A) + R_x(x)  = J(A, x) \label{eq:MAP_blind}
 \end{align}
 We used the Bayes rule to go from the first to the second line and applied the function $-\log$ to get to the third.
 This formulation often appears in the literature and is at the basis of the most successful handcrafted approaches, see e.g. the review paper \cite{campisi2017blind}.

Let $\hat x(A,y)$ denote the minimizer of \eqref{eq:MAP_blind} with $A$ fixed. 
Injecting this into the cost function, we see that finding the optimal operator $A$ in the problem above is equivalent to:
 \begin{equation*}
    \argmin_{A\in \K^{M\times N}} \frac{1}{2\sigma^2} \|A\hat x(A,y)-y\|_2^2 + R_A(A) + R_x(\hat x(A,y)).
\end{equation*}
Assuming that the distributions $\mu_x$ and $\mu_A$ are uniform over compact sets, the functions $R_A$ and $R_x$ are constant. 
Hence, the MAP approach finally simplifies to:
 \begin{equation*}
    \min_{\theta\in \Theta} \frac{1}{2\sigma^2} \|A(\theta)\hat x(A(\theta),y)-y\|_2^2,
\end{equation*}
which coincides with the proposed approach. 

\paragraph{A type of P\&P}

Let us mention that the proposed approach can also be seen as a type of P\&P prior. 
Instead of training a denoising network, as is the case in the initial P\&P approach, we train an inverse problem solver. 
We then use it as a prior to infer an operator, instead of a signal.
Making deeper links with this approach is out of the scope of this paper.}

\subsection{Numerical resolution\label{sec:numerical_bip}}

As the function in \eqref{eq:blindinvmintheta} is deterministic over a small, to moderate dimension (between 5 and 1000 for the considered applications), we can opt for many different $0$-th or $1$-st order optimization routines.

Applying $1$-st order methods is highly non trivial without using automatic differentiation.
Indeed, we need to compute the Jacobian of $\Nc\left(\wz^\star, A(\thetaz), \yz\right)$ with respect to the parameter $\thetaz$. 
This in particular requires evaluating the derivative of $A(\thetaz)$ with respect to the parameters $\thetaz$ and of the neural network $\Nc\left(\wz^\star, A(\thetaz), \yz\right)$ with respect to its second variable. In all our experiments, we used the automatic differentiation techniques available in PyTorch. \rev{This required to implement the Jacobian of the mapping $A$ with respect to $\theta$. To actually solve the problem, we considered the following optimization routines:}
\begin{itemize}
    \item The L-BFGS optimizer \cite{liu1989limited}. This quasi-Newton method estimates the Hessian of the function using first order information only and is known to converge rapidly when initialized close to a (local) minimizer. It therefore seems particularly adapted when the user has a good knowledge of the true parameter $\bar \thetaz$. \rev{We used this approach for the MRI experiments.}
    \item The RMSProp or ADAM optimizer \cite{tieleman2012divide,kingma2014adam}.
    \rev{For the computerized tomography experiments, we observed issues with a convergence to bad local minimizers using L-BFGS}. To avoid this phenomenon, a possibility is to resort to inertial methods, which are known to escape narrow basins of attraction. In our experiments, we used the RMSProp optimizer with a parameter $\beta=0.9$ (Adam with $\beta_1=0$). \rev{This procedure turned out to provide satisfactory results consistently.}
    \item Bayesian optimization \cite{frazier2018bayesian}. \rev{In some cases, it can be helpul to resort to $0$-th order methods. This is the case if the mapping $A(\theta)$ is not differentiable with respect to the parameter $\theta$. This is also the case if the cost function is too chaotic in which case, the gradient of the objective function does not provide a meaningful information on the location of the global minimizer. We can then resort to Bayesian optimization techniques. They typically work reliably for moderate dimensions $1-20$. We used this approach for the deblurring experiments, since it only involves 7 parameters and that we wanted to secure finding a good approximation of the global minimizer.}
\end{itemize}

\section{Numerical experiments}
\label{sec:results}

The numerical experiments are divided in two sections. In the first section \ref{sec:family_therapy} we compare the benefits and drawbacks of training model-based networks on a family of operators. 

In the second section \ref{sec:blind:xp}, we illustrate that training model-based networks on a family of operators allows solving blind inverse problems.
The experiments are carefully conducted on the three applications: MRI, CT and image deblurring with an unrolled ADMM \cite{sun2016deep}.

\subsection{\rev{Training} setting}

All the models were trained using the Adam optimizer in PyTorch with the default parameters except the learning rate which was tuned for each experiment. \rev{We observed that depending on the imaging modality, the default learning rate could lead to a divergent sequence. In those situations, we divided it by 10 until we observed empirical convergence. The basic idea is to obtain a sufficient decay rate at the first epoch, without diverging.}

\rev{\paragraph{Denoising}

For the P\&P experiments, we trained the DRUNet model of \cite{zhang2021plug} for grayscale images using the ImageNet database.
We trained it for 100 epochs.
We used a number of channels equal to 32, 64, 128, 256 for the 4 different layers. 
This results in a lighter network than the initial one, where the number of channels was doubled for each layer: 64, 128, 256, 512.
The reason for this choice is to get a lighter model compatible with the unrolled architectures.
}

\paragraph{Magnetic Resonance Imaging}

The training database is the fastMRI knee training dataset \cite{zbontar2018fastmri}. It contains $34,742$ images of size $320\times 320$.
All evaluations were performed on the validation set of the fastMRI knee database containing $7,135$ 2D slices.
We used the efficient cuFINUFFT transform \cite{shih2021cufinufft}, which is the fastest available library in our experiments (see \url{https://github.com/albangossard/Bindings-NUFFT-pytorch} for comparisons).

For the experiments illustrating the advantages of training on a family of operators, we set $M=N/4$, i.e. a 4x downsampling rate. We used a single reception coil ($J=1$) with a known sensitivity map $s=1$. 
The denoising network $\Nc^d$ was trained on $30$ epochs with a learning rate of $10^{-3}$ and an exponential step decay of $0.95$ after each epoch.
The unrolled network $\Nc^a$ uses $K=10$ iterations and it was trained on $14$ epochs with a learning rate of $10^{-4}$ and an exponential step decay of $0.95$ after each epoch.
Both training took about 24h on an Nvidia V100, resulting in a total energy consumption of $\sim 70$kWh.

The blind reconstruction experiments are conducted with $M=N/10$ measurements and $J=15$ reception coils.
The networks are trained for $8$ epochs with a learning rate of $10^{-4}$ and with an exponential step decay of $0.95$ after each epoch.

\rev{These subsampling rates are used frequently in MRI experiments.}

\paragraph{Computerized Tomography}

\rev{We trained the network using $K=5$ iterations using the ImageNet database. We initially used the Lung Image Database Consortium \cite{armato2011lung} database, but realized that it contains many improper slices (high noise, streaking artifacts, little contents...). We evaluated the algorithm on a curated version called LoPoDaP \cite{Leuschner2021}, containing less artifacts. The test dataset contains 4096 images.

As the blind inverse problem \eqref{eq:blindinvmintheta} requires differentiating the operator $\Az(\thetaz)$ with respect to its parameters $\thetaz$, we cannot use standard GPU-based libraries to compute the Radon transform \cite{ronchetti2020torchradon}.
We thus resorted to an homemade implementation that relies on a NUFT through the Fourier slice theorem.
In order to reduce the important numerical cost and energy consumption of the experiments with CT reconstruction, we downsized the images to $256\times 256$.
}

\paragraph{Deblurring}

The image deblurring experiments were carried out with the MS COCO dataset \cite{lin2014microsoft} ($118,287/5,000$ images for training/validation). During training we randomly cropped patches of size $400\times 400$ to speed-up the computation.

\subsection{Benefits of training on a family}\label{sec:family_therapy}

\subsubsection{Training on fixed operators}\label{sec:lack_adaptivity}

In this section, we highlight the limits of training a reconstruction network on a single operator, as is currently the dominant practice. Let us detail the training procedure for each application. 

For MRI reconstruction, we considered measurements coming from a single reception coil, to reduce the computational complexity. We used both the denoising network $\Nc^d$ and the unrolled proximal gradient descent $\Nc^p$ on $5$ different schemes: a radial one ($\bigcirc$, Fig.~\ref{fig:ref}), a horizontal one ($-$, Fig.~\ref{fig:hor}) and a vertical one ($|$, Fig.~\ref{fig:vert}). In addition, we used two crosses, which do not belong to the training family $\Ac$. The first one is aligned with the axes ($+$, Fig.~\ref{fig:axes}) and the other one with the diagonals ($\times$, Fig.~\ref{fig:diag}).

For CT reconstruction, we considered measurements coming from randomly perturbed versions $\thetaz_1,\thetaz_2,\thetaz_3$ of the equiangular pattern $\thetaz_0$. The network is an unrolled ADMM $\Nc^a$ ran for $4$ iterations. The perturbations $\thetaz_1,\thetaz_2$ belong to the family $\Ac$ used for the family training. The perturbation $\thetaz_3$ is twice larger than what was observed during the training phase and does not belong to $\Ac$.

For the deblurring problem, we considered random convolution kernels generated using the model \eqref{eq:zernike_parametrization}. The network is also an unrolled ADMM $\Nc^a$ ran for $K=5$ iterations. 
The perturbations $\theta_0, \thetaz_1,\thetaz_2$ belong to the family $\Ac$ used for the family training, \rev{while $\theta_3, \theta_4$ do not and have a larger spatial spread.}

In Table~\ref{tab:single_table}, we report the average peak signal-to-noise ratio on the validation set.
Table~\ref{tab:single_table} illustrates various observations listed below.

\paragraph{Lack of adaptivity} 
The values on the diagonal are higher than the off-diagonal terms, \rev{except for the CT experiment where the family trained network is slightly better in average}.
This just reflects the fact that the best way to reconstruct images for a given application is to train the network for this specific application. 

\begin{figure*}[h]
    \centering
    \begin{subfigure}[t]{\imagewidth}
        \centering
        \plotwithzoom{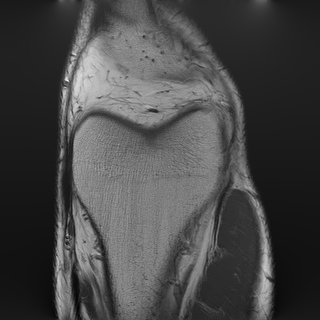}
        \vspace{-1.5em}
        \captionsetup{justification=centering}
        \caption{Original image}
        \label{fig:img_recon:true}
    \end{subfigure}
    \begin{subfigure}[t]{\imagewidth}
        \centering
        \plotwithzoom{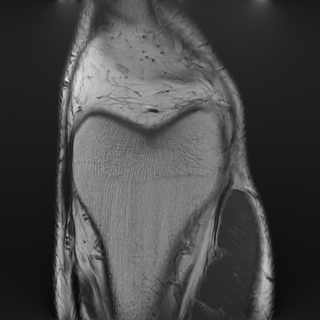}
        \vspace{-1.5em}
        \captionsetup{justification=centering}
        \caption{Tr.$|$ Ev.$|$, $38.9$dB}
        \label{fig:img_recon:truexi}
    \end{subfigure}\\
    \begin{subfigure}[t]{\imagewidth}
        \centering
        \plotwithzoom{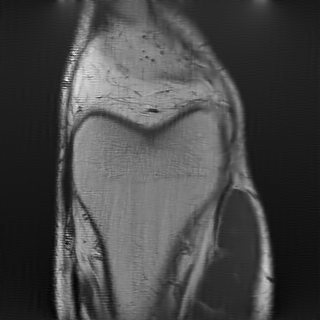}
        \vspace{-1.5em}
        \captionsetup{justification=centering}
        \caption{Tr.$|$ Ev.$-$, $33.7$dB}
        \label{fig:img_recon:otherxi}
    \end{subfigure}
    \begin{subfigure}[t]{\imagewidth}
        \centering
        \plotwithzoom{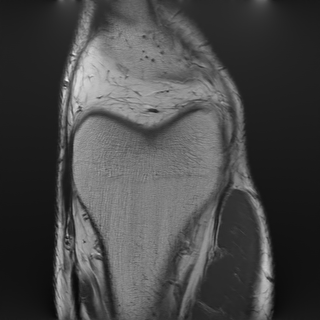}
        \vspace{-1.5em}
        \captionsetup{justification=centering}
        \caption{FU Ev.$-$, $38.0$dB}
        \label{fig:img_recon:family}
    \end{subfigure}
    \caption{Examples of reconstructions using the MRI unrolled network. We trained it on a single vertical $|$ sampling pattern in Fig. \ref{fig:img_recon:truexi}, \ref{fig:img_recon:otherxi} and on a family in Fig. \ref{fig:img_recon:family}.
    We tested it on the vertical $|$ sampling pattern in Fig. \ref{fig:img_recon:truexi} and on the horizontal $-$ pattern in Fig. \ref{fig:img_recon:otherxi}, \ref{fig:img_recon:family}. Observe the huge gain in adaptivity when training on a family.}
    \label{fig:img_recon}
\end{figure*}

The drop of peak signal-to-noise ratio (PSNR) when using a network trained with the wrong operator can be as high as $9$dB for the denoising net on the MRI experiment (see trained on $|$, applied on $-$). This drop is more moderate, but yet really significant (MRI: $5$dB, \rev{CT: $2.4$dB}, Blur: $22$dB) for the unrolled net. This is a striking illustration of the strong dependency of a reconstruction network to the operator used at the training stage. We illustrate the artifacts that can appear when the operator is trained on a different operator for the MRI application in Fig.~\ref{fig:img_recon}. We can clearly see horizontal stripes oscillating at a high frequency, suggesting that the network did not properly learn to reconstruct the corresponding Fourier coefficients.

\paragraph{Peculiar case of deblurring and CT nets}

\rev{The deblurring application has an important peculiarity: the basic block of the convolutional neural network is identical to the forward operator. Hence, when an unrolled network is trained, we can expect the networks $\Dc(\wz_k,\cdot)$ to not only act as ``denoisers'', but also as deconvolution mappings.
This fact might explain the catastrophic lack of adaptivity in Table \ref{fig:tab_fixed:blur}. For instance with the network trained on $\theta_2$, we obtain an average performance of $27.5$dB without training mismatch and less than $10$dB with a mismatch. This also confirms the conclusions of the introductory example in Fig. \ref{tab:exampleissues}.

A similar, yet less obvious phenomenon seems to occur with the CT experiment. When training a network on the equiangular pattern $\theta_0$, the lack of adaptivity is particularly striking. We believe that this may as well be due to a particular ``algebraic compatibility'' between convolution operators and the regularly spaced Radon transform. Further investigations should be conducted to further strengthen this hypothesis.}

\paragraph{Superiority of unrolled nets} The unrolled networks provide better reconstruction results than the denoising net in the MRI experiment. The overall gain on the diagonal varies between $1.4$dB and $1.7$dB for this particular application, which is significant. This is in accordance with recent comparisons of both strategies \cite{muckley2021results}. Hence, we only consider unrolled nets for the forthcoming experiments.

\paragraph{Optimal acquisition schemes} 
Looking at the diagonal of the tables in Table \ref{tab:single_table} reveals that some acquisition schemes are better than others when the networks are trained properly. In the MRI experiment for instance, we see that the $+$ sampling scheme, yields a PSNR of $38.0$dB in average while it drops to $37.1$dB for the $-$ sampling scheme. This is in accordance with recent results \cite{wang2022b,weiss2021pilot,gossard2022bayesian}. 
For CT, the best sampling scheme is the standard equispaced one, which may not come as a surprise. 
In deblurring, it seems that the blur related to $\thetaz_3$ is particularly hard to invert. This is probably due to a larger spatial spread.

\begin{table*}
    \centering
    \begin{subfigure}[b]{\tabwidth}
        \centering
        \includegraphics[width=\textwidth]{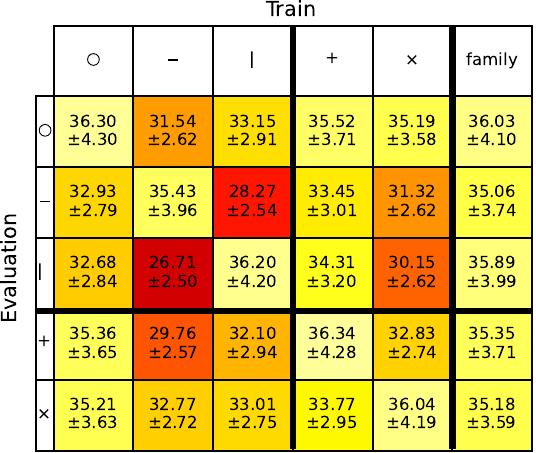}
        \caption{MRI: denoising net}
        \label{fig:tab_fixed:den}
    \end{subfigure}
    \begin{subfigure}[b]{\tabwidth}
        \centering
        \includegraphics[width=\textwidth]{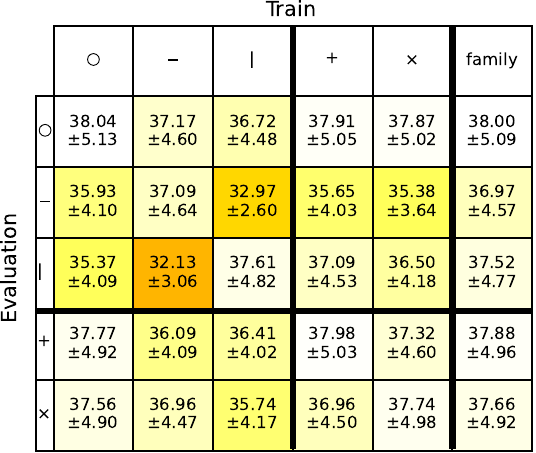}
        \caption{MRI: unrolled net}
        \label{fig:tab_fixed:fb}
    \end{subfigure}
    \begin{subfigure}[b]{\tabwidth}
        \centering
        \includegraphics[width=\textwidth]{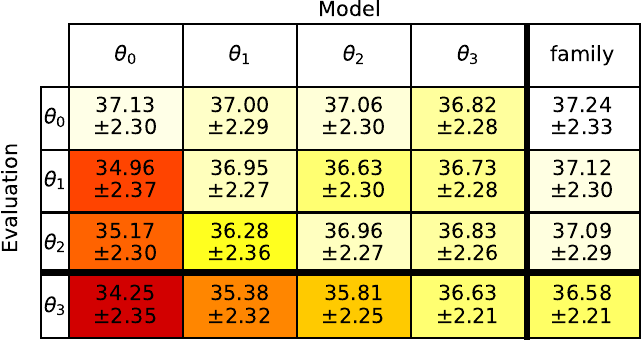}
        \caption{CT: unrolled net}
        \label{fig:tab_fixed:CT}
    \end{subfigure}
    \begin{subfigure}[b]{\tabwidth}
        \centering
        \includegraphics[width=\textwidth]{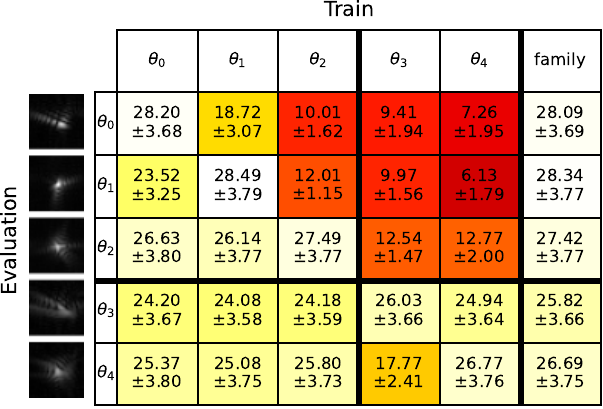}
        \caption{Deblurring: unrolled net}
        \label{fig:tab_fixed:blur}
    \end{subfigure}
    \caption{The lack of adaptivity. In these tables, we measure the performance of various solvers for non blind inverse problems. We train a network for a given operator and test it on others. The results for a network trained on a family of forward operators is also given in the last column of each table. The average PSNR and its standard deviation are evaluated on the different validation dataset. \rev{This means on about 7 000 images in MRI, 4 096 in CT and 5 000 in deblurring.}}
    \label{tab:single_table}
\end{table*}

\subsubsection{Training on an operator family}\label{sec:training_family}

\rev{Let us now study what happens, when training} the reconstruction networks by varying the forward operators, as in \eqref{eq:main_idea_of_the paper}. 
In what follows, we let ID denote the \emph{``ideal'' denoising} network and IU denote the \emph{``ideal'' unrolled} network. 
By ideal, we mean that the networks have been trained and tested with the same operator. They serve as a benchmark \rev{that cannot be outperformed on the training dataset for a given architecture}.
We let FD and FU denote the \emph{``family'' denoising} and \emph{family unrolled}  networks, which have been trained over a complete family.
We also tested the P\&P approach. We used an unrolled ADMM for different numbers of iterations $K$. The state-of-the-art DRUNet \cite{zhang2021plug} network was used as an embedded denoiser \rev{and it was carefully trained on the FastMRI dataset for MRI, on MS COCO for deblurring and on ImageNet for the CT experiment (since the LIDC \cite{armato2011lung} and LoDoBap \cite{Leuschner2021} datasets contain images with many artifacts).}
It was trained specifically to denoise the images with various levels of white Gaussian noise. 
Finally, we implemented the reuse \& regularize network (R\&R) \cite{gilton2021model} composed of $K=10$ iterations.
The embedded inversion network consists of a pseudo-inverse, followed by a DRUNet network trained for the $\bigcirc$ sampling scheme. The hyperparameters in the method (see \cite{gilton2021model}) were tuned to produce the best results. 
Table~\ref{tab:family_table} shows the performance of the different architectures. The following conclusions can be drawn.

\begin{table*}
    \centering
    \begin{subfigure}[b]{1\textwidth}
        \includegraphics[width=1\textwidth]{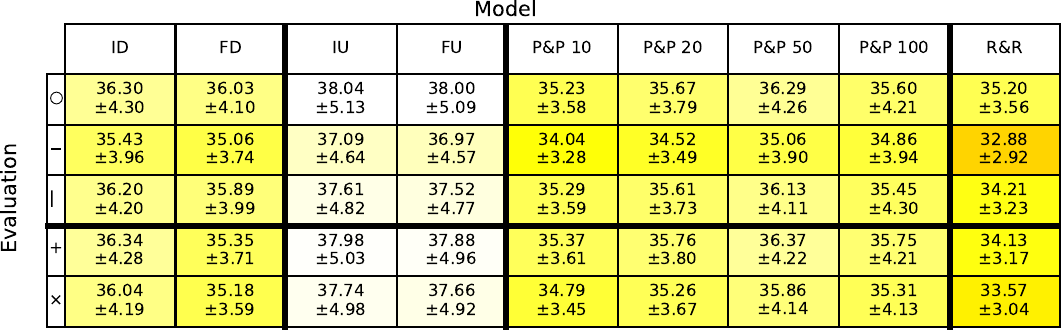}
        \caption{MRI -- denoising (ID, FD) and unrolled (IU, FU)}
    \end{subfigure}
    \begin{subfigure}[b]{0.45\textwidth}
        \includegraphics[width=1\textwidth]{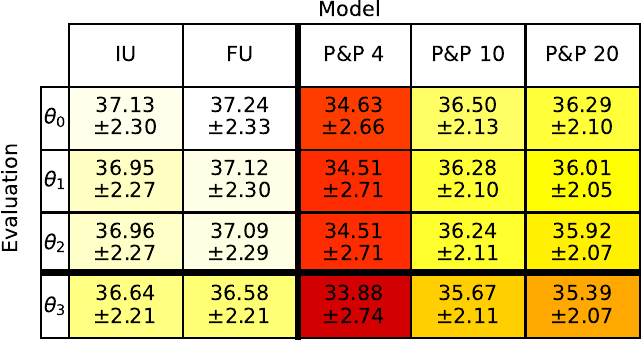}
        \caption{CT -- unrolled}
    \end{subfigure}
    \begin{subfigure}[b]{0.45\textwidth}
        \includegraphics[width=1\textwidth]{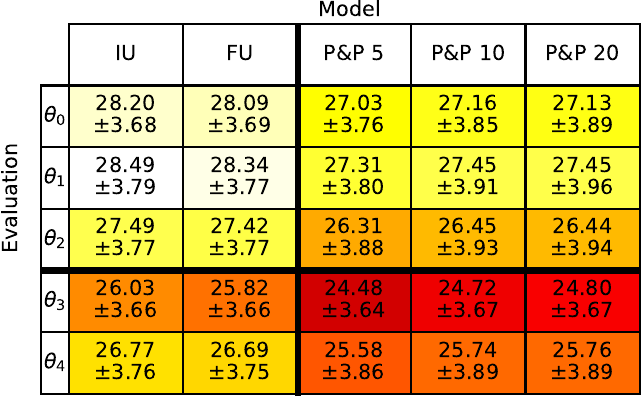}
        \caption{Deblurring -- unrolled}
    \end{subfigure}
    \caption{The average PSNR and standard deviation in dB, for various reconstruction approaches and operators. 
    \rev{The evaluation dataset contains about 7 000 images in MRI, 4 096 in CT and 5 000 in deblurring.}
    ID, IU (``ideal''): the network is trained on the same operator it is applied on. FD, FU (``family''): the network is trained on a family of operators, as advocated in this paper. P\&P: plug-and-play ADMM network with different numbers of iterations $K \in \{4,5,10,20,50,100\}$. Notice that the schemes $+,\times$ for MRI and the parameters $\thetaz_3, \thetaz_4$ for CT and image deblurring do not belong to the training family and therefore allow assessing the generalization capability of the reconstruction networks.}
    \label{tab:family_table}
\end{table*}

\paragraph{Price of adaptivity}

By comparing the columns FU and IU in Table \ref{tab:family_table}, we see that training on a family leads to really moderate drops of performance compared to a training on a single operator. \rev{Surprisingly, it even outperforms the model trained on a single operator for the CT experiment. This might be caused by a better capacity to escape spurious minimizers at the training stage, or by a discrepancy between the testing and training datasets.}
Some differences are still significant for the denoising network, but they only become marginal for the unrolled networks. This has to be compared to the huge gain of adaptivity of the method: a single network is now able to tackle a vast family of different problems within a class.

For operators in the training family, the performance drop of unrolled networks is of at most $0.12$dB in MRI ($-$), and $0.15$dB in deblurring ($\theta_1$) and there is no drop for CT. Compared to values reaching more that $10$dB in the previous section, this feature is really remarkable. This perfectly illustrates one of the take home message of our paper: training unrolled networks on a family seems to not degrade the performance significantly while providing a huge boost of adaptivity.

\paragraph{How does it generalize?}

It is informative to look at the last rows of the different tables in Table \ref{tab:family_table}. There, we apply the unrolled networks to operators that were not encountered at the training stage. Hence, comparing IU to FU allows us to assess the generalization ability of the networks.
We observe a performance drop of \rev{$0.1$dB at most in MRI, $0.06$dB in CT and $0.21$dB in deblurring.}
It can also be counterbalanced by the fact that the operators differ significantly from what was observed at the training stage: we amplified the perturbations by a factor $2$ for the CT and deblurring experiments. Overall, this experiment suggests that a training stage on a family provides some generalization capabilities.

\paragraph{Plug \& Play (P\&P)}

\rev{When looking at Table~\ref{tab:family_table}, we see that the P\&P approach is outperformed uniformly by both unrolled networks trained on a family and on a fixed operator.
The drop lies between $1$ and $2$dB for the MRI experiments, about $1$dB for the deblurring experiments and less than $1$dB for the CT experiments.
This suggests that for a given reconstruction architecture, it is beneficial to train the proximal networks for a specific task rather than using a \emph{universal} denoiser, as is the case in P\&P. \rev{This observation should be carefully examined with recent progress in diffusion models \cite{graikos2022diffusion}.}
Notice however, that compared to the off-diagonal elements of Table~\ref{tab:single_table} which correspond to a network trained on a fixed operator and evaluated with a different one, the P\&P approach is still really competitive and likely preferable.}

\rev{We also want to mention that FU} does not seem to extrapolate well to problems completely different from the ones it was trained for. 
Indeed, we trained FU for an MRI reconstruction problem and tested it for a deblurring application. 
In this application, which is not reported in this paper, the P\&P approach was considerably more consistent. 
In a sense, we can see the proposed training as an intermediate step between the P\&P approach (adaptable to all inverse problems) and the traditional training of reconstruction networks (perfectly adapted to a single operator).

\paragraph{Reuse \& Regularize (R\&R)} 
Finally, the R\&R approach applied in MRI can improve the results for some problems compared to a model trained on a single operator. However, it seems that our simpler training approach provides significantly better results. Hence, we did not consider this alternative for the CT and deblurring problems.

\subsection{Blind inverse problems}\label{sec:blind:xp}

In this section, we illustrate how training on a family of operators helps solving different blind inverse problems. 
We assume that 
\begin{equation}
    \yz = \Az(\thetaz_1) \xz +\bz,
\end{equation}
for some unknown parameter $\thetaz_1$ describing the forward model. 
We then solve~\eqref{eq:blindinvmintheta} \rev{using the methods described in Section \ref{sec:numerical_bip}}, resulting in an estimate $\hat \thetaz_1$ of $\thetaz_1$. Fig. \ref{tab:blind:mri}, \ref{tab:blind:ct}, \ref{tab:blind:blur} show the performance of the solver for various applications. Let us analyze these results.

\subsubsection{Magnetic Resonance Imaging}

\begin{figure*}[h]
    \def\subfigwidth{0.24\textwidth}
    \def\trimzoomleft{.3}
    \def\trimzoomlower{.6}
    \def\trimzoomright{.55}
    \def\trimzoomupper{.25}
    \centering
    \footnotesize
    \begin{tabular}{@{}cc@{}c@{}c@{}c@{}}
         &
        \makecell{
        \begin{tikzpicture}
            \node[inner sep=0pt] (img) at (0,0) {
                \includegraphics[width=\subfigwidth]{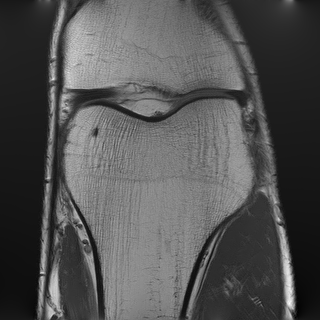}%
                \llap{\adjincludegraphics[width=0.1\textwidth,trim={{\trimzoomleft\width} {\trimzoomlower\height} {\trimzoomright\width} {\trimzoomupper\height}},clip,cfbox=red 0.5pt 0pt]{f_Nlines=16_idimg=20.png}}
            };
            \draw [stealth-,red] (-0.4,0.35) -- (0.27,-0.25);
        \end{tikzpicture}
        \\ Ground truth} &
        \makecell{\includegraphics[width=\subfigwidth]{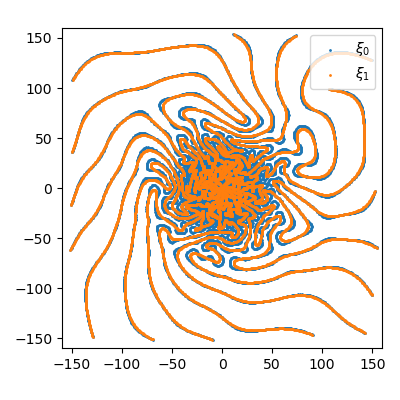}\\ Measured traj. $\xi_1$\\ and target traj. $\xi_0$} &
        \makecell{\includegraphics[width=\subfigwidth]{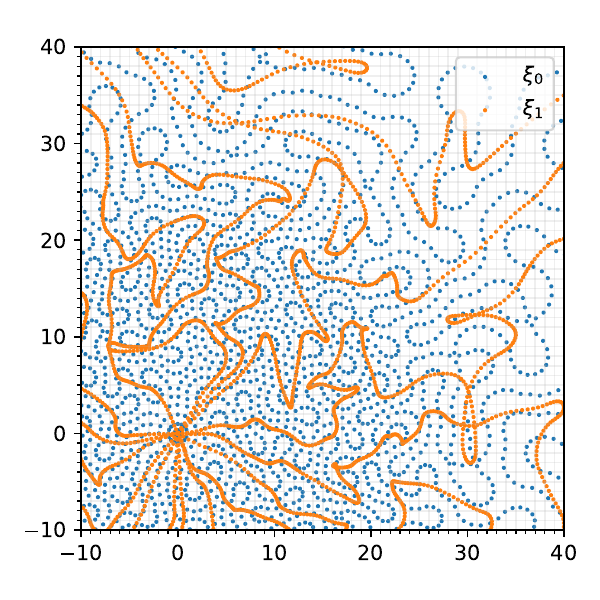}\\$k$-space zoom\\
        $\Vert\xi_1-\xi_0\Vert_\infty=5.01$\\
        $PSNR(s_0,s_1)=25.39$dB} \\
        \midrule

        \rotatebox[origin=c]{90}{Trained on $\thetaz_0$} &
        \makecell{
        \begin{tikzpicture}
            \node[inner sep=0pt] (img) at (0,0) {
                \includegraphics[width=\subfigwidth]{f_tilde_Nlines=16_idimg=20_nonblind_fixed.png}%
                \llap{\adjincludegraphics[width=0.1\textwidth,trim={{\trimzoomleft\width} {\trimzoomlower\height} {\trimzoomright\width} {\trimzoomupper\height}},clip,cfbox=red 0.5pt 0pt]{f_tilde_Nlines=16_idimg=20_nonblind_fixed.png}}
            };
            \draw [stealth-,red] (-0.4,0.35) -- (0.27,-0.25);
        \end{tikzpicture}
        \\ $30.63$dB} &
        \makecell{
        \begin{tikzpicture}
            \node[inner sep=0pt] (img) at (0,0) {
                \includegraphics[width=\subfigwidth]{f_tilde0_Nlines=16_idimg=20_fixed.png}%
                \llap{\adjincludegraphics[width=0.1\textwidth,trim={{\trimzoomleft\width} {\trimzoomlower\height} {\trimzoomright\width} {\trimzoomupper\height}},clip,cfbox=red 0.5pt 0pt]{f_tilde0_Nlines=16_idimg=20_fixed.png}}
            };
            \draw [stealth-,red] (-0.4,0.35) -- (0.27,-0.25);
        \end{tikzpicture}
        \\ $15.17$dB} &
        \makecell{
        \begin{tikzpicture}
            \node[inner sep=0pt] (img) at (0,0) {
                \includegraphics[width=\subfigwidth]{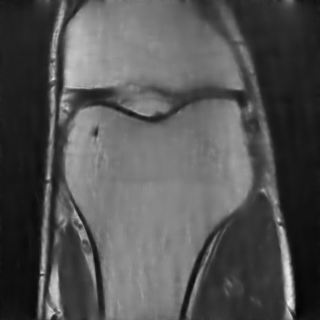}%
                \llap{\adjincludegraphics[width=0.1\textwidth,trim={{\trimzoomleft\width} {\trimzoomlower\height} {\trimzoomright\width} {\trimzoomupper\height}},clip,cfbox=red 0.5pt 0pt]{f_tilde_Nlines=16_idimg=20_fixed.png}}
            };
            \draw [stealth-,red] (-0.4,0.35) -- (0.27,-0.25);
        \end{tikzpicture}
        \\ $30.63$dB} &
        \makecell{\includegraphics[width=\subfigwidth]{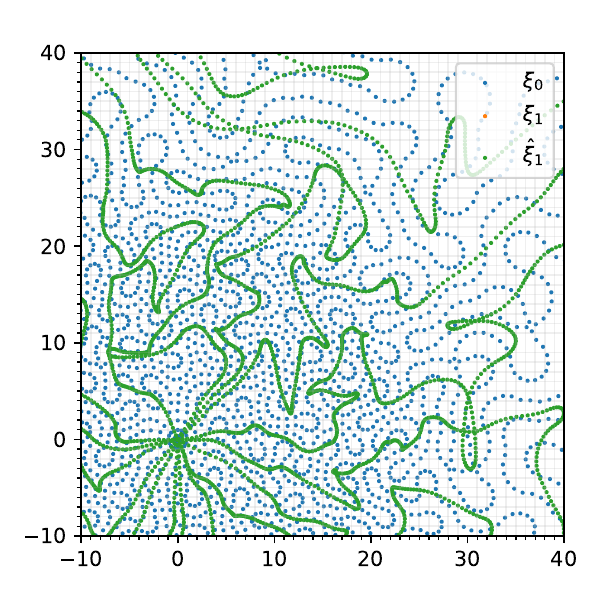}\\
        $\Vert\xi_1-\hat\xi_1\Vert_\infty=0.0082$\\
        $PSNR(\hat s_1, s_1)=65.84$dB} \\

        \rotatebox[origin=c]{90}{Trained on a family} &
        \makecell{
        \begin{tikzpicture}
            \node[inner sep=0pt] (img) at (0,0) {
                \includegraphics[width=\subfigwidth]{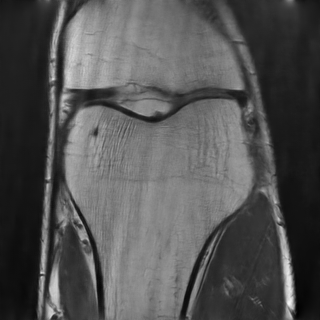}%
                \llap{\adjincludegraphics[width=0.1\textwidth,trim={{\trimzoomleft\width} {\trimzoomlower\height} {\trimzoomright\width} {\trimzoomupper\height}},clip,cfbox=red 0.5pt 0pt]{f_tilde_Nlines=16_idimg=20_nonblind_family.png}}
            };
            \draw [stealth-,red] (-0.4,0.35) -- (0.27,-0.25);
        \end{tikzpicture}
        \\ $31.66$dB} &
        \makecell{
        \begin{tikzpicture}
            \node[inner sep=0pt] (img) at (0,0) {
                \includegraphics[width=\subfigwidth]{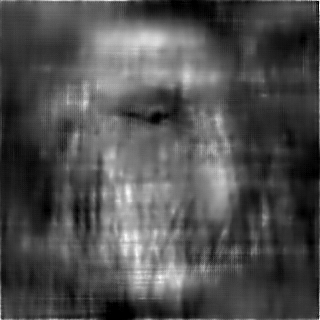}%
                \llap{\adjincludegraphics[width=0.1\textwidth,trim={{\trimzoomleft\width} {\trimzoomlower\height} {\trimzoomright\width} {\trimzoomupper\height}},clip,cfbox=red 0.5pt 0pt]{f_tilde0_Nlines=16_idimg=20_family.png}}
            };
            \draw [stealth-,red] (-0.4,0.35) -- (0.27,-0.25);
        \end{tikzpicture}
        \\ $15.35$dB} &
        \makecell{
        \begin{tikzpicture}
            \node[inner sep=0pt] (img) at (0,0) {
                \includegraphics[width=\subfigwidth]{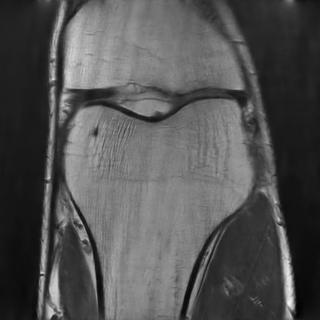}%
                \llap{\adjincludegraphics[width=0.1\textwidth,trim={{\trimzoomleft\width} {\trimzoomlower\height} {\trimzoomright\width} {\trimzoomupper\height}},clip,cfbox=red 0.5pt 0pt]{f_tilde_Nlines=16_idimg=20_family.png}}
            };
            \draw [stealth-,red] (-0.4,0.35) -- (0.27,-0.25);
        \end{tikzpicture}
        \\ $32.00$dB} &
        \makecell{\includegraphics[width=\subfigwidth]{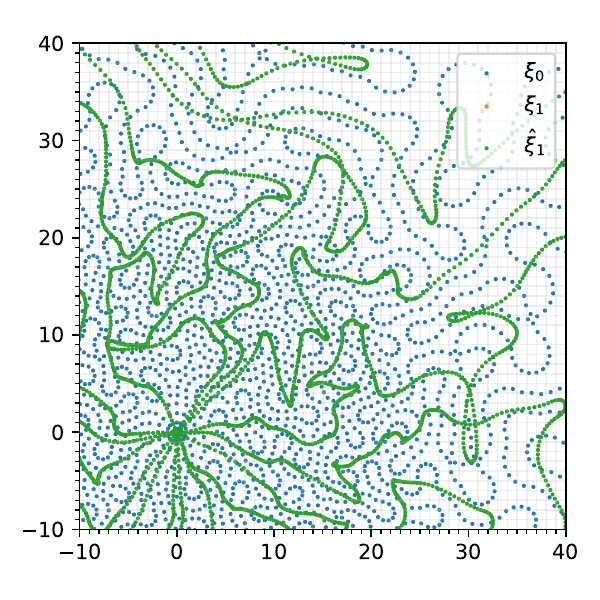}\\
        $\Vert\xi_1-\hat\xi_1\Vert_\infty=0.034$\\
        $PSNR(\hat s_1, s_1)=47.35$dB} \\

        \rotatebox[origin=c]{90}{P\&P} &
        \makecell{
        \begin{tikzpicture}
            \node[inner sep=0pt] (img) at (0,0) {
                \includegraphics[width=\subfigwidth]{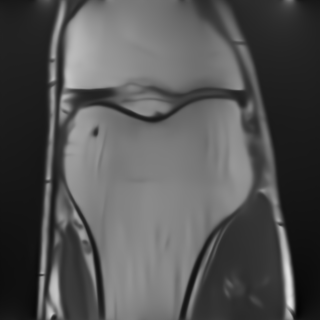}%
                \llap{\adjincludegraphics[width=0.1\textwidth,trim={{\trimzoomleft\width} {\trimzoomlower\height} {\trimzoomright\width} {\trimzoomupper\height}},clip,cfbox=red 0.5pt 0pt]{f_tilde_Nlines=16_idimg=20_nonblind_PP.png}}
            };
            \draw [stealth-,red] (-0.4,0.35) -- (0.27,-0.25);
        \end{tikzpicture}
        \\ $30.50$dB} &
        \makecell{
        \begin{tikzpicture}
            \node[inner sep=0pt] (img) at (0,0) {
                \includegraphics[width=\subfigwidth]{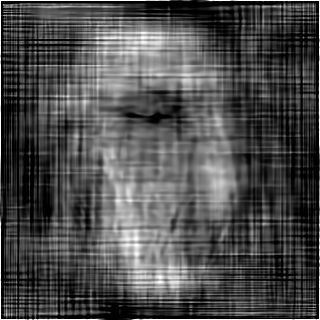}%
                \llap{\adjincludegraphics[width=0.1\textwidth,trim={{\trimzoomleft\width} {\trimzoomlower\height} {\trimzoomright\width} {\trimzoomupper\height}},clip,cfbox=red 0.5pt 0pt]{f_tilde0_Nlines=16_idimg=20_PP.png}}
            };
            \draw [stealth-,red] (-0.4,0.35) -- (0.27,-0.25);
        \end{tikzpicture}
        \\ $12.63$dB} &
        \makecell{
        \begin{tikzpicture}
            \node[inner sep=0pt] (img) at (0,0) {
                \includegraphics[width=\subfigwidth]{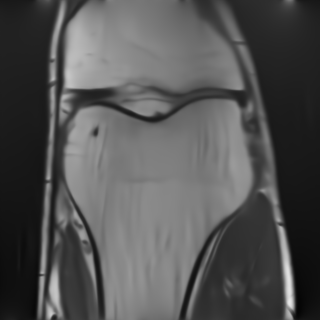}%
                \llap{\adjincludegraphics[width=0.1\textwidth,trim={{\trimzoomleft\width} {\trimzoomlower\height} {\trimzoomright\width} {\trimzoomupper\height}},clip,cfbox=red 0.5pt 0pt]{f_tilde_Nlines=16_idimg=20_PP.png}}
            };
            \draw [stealth-,red] (-0.4,0.35) -- (0.27,-0.25);
        \end{tikzpicture}
        \\ $30.50$dB} &
        \makecell{\includegraphics[width=\subfigwidth]{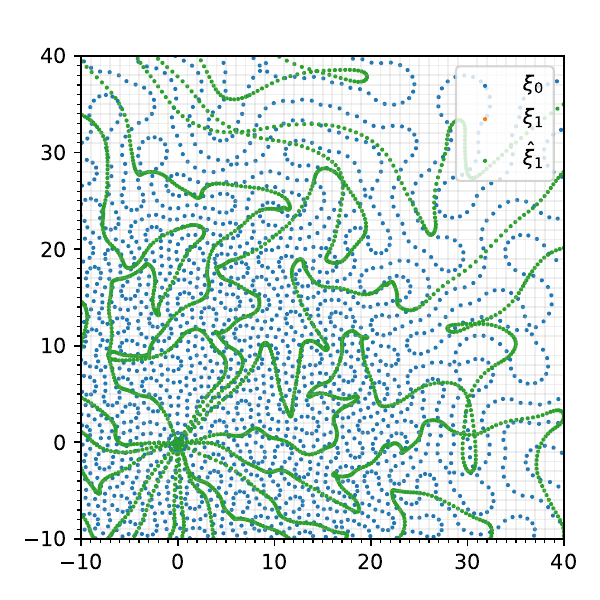}\\
        $\Vert\xi_1-\hat\xi_1\Vert_\infty=0.0039$\\
        $PSNR(\hat s_1,s_1)=69.71$dB} \\

         &
        $\Nc^a(\wz, \thetaz_1, \yz)$ &
        $\Nc^a(\wz, \thetaz_0, \yz)$ &
        $\Nc^a(\wz, \hat\thetaz_1, \yz)$ &
        Trajectories $\xi$ \\
    \end{tabular}
    \caption{Self-calibrated MRI.
    \emph{1st column:} reconstruction with a perfect knowledge of the forward model $\thetaz_1$.
    \emph{2nd column:} reconstruction assuming the wrong forward model $\thetaz_0$.
    \emph{3rd column:} reconstruction using the estimated forward model $ \hat\thetaz_1$.
    \emph{4th column:} estimate of the operator. We display the maximal distance between sampling points $\|\xi_1-\hat \xi_1\|_\infty$ as well as the PSNR of the estimated sensitivity maps $\hat s_1$.
    From top to bottom: different training strategies are compared.
    \emph{2nd row:} trained on $\thetaz_0$.
    \emph{3rd row:} trained on a family of operators.
    \emph{4th row:} using a P\&P prior.
    The PSNR is indicated below each image. The noise level given to the P\&P denoiser has been tuned so as to yield the best PSNR on the non-blind problem. Other models do not require tuning at evaluation.}
    \label{tab:blind:mri}
\end{figure*}

This application provides surprisingly good results for various reasons:
\begin{itemize}
    \item To the best of our knowledge, no one yet attempted to estimate the sensitivity maps and trajectory errors jointly. Estimating divergence in trajectories might look hopeless at first sight, which may explain this fact. Indeed, looking at the differences between $\xi_1$ and $\xi_0$ (see top-right and the zoom on the right-most column of Fig. \ref{tab:blind:mri}) we see that the frequency shifts are huge (up to 5 pixels). 
    \item The total number of parameters to estimate is large. Indeed, it consists in the $104\times 15$ parameters describing the sensitivity maps and the $32$ parameters describing the convolution kernel that perturbs the trajectories, i.e. $1592$ parameters.
\end{itemize}

If solved without any correction, the reconstruction results are disastrous (see the 2nd column).
Solving the consistency problem \eqref{eq:blindinvmintheta} provides near perfect estimates of $\hat \thetaz$ for all reconstruction mappings. For instance, the green $\hat \xi_1$ and orange $\xi_1$ trajectories cannot be distinguished on the right column. 
This may come as a surprise, and seems to suggest that this particular blind inverse problem is not as hard as it may seem at first sight. This might be due to some redundancy in the data: the $15$ reception coils associated to a slight oversampling of the $k$-space center (all the trajectories start exactly from the center) seem to ensure the identifiability of the problem. 
A nice research perspective is to explain this phenomenon from a theoretical viewpoint.

The reconstruction result obtained with the neural network trained on a family is significantly better than the two other ones (more than $+1.3$dB compared to the one trained on $\thetaz_0$ and to the P\&P approach). In particular, the bone texture is reconstructed with the proposed approach, while it is not for the two others.

\rev{
To further validate the method, we tested the methodology on 9 additional images. The results are reported in Table \ref{tab:blind_MRI}. 
As can be seen, the method recovers good estimates of the sensitivity maps $s_1$ and trajectories $\xi_1$ in all cases. 
This results in a huge PSNR increase, since the forward model is essentially correct after estimation. 
\begin{table*}
    \footnotesize
    \centering
    \begin{tabular}{c|c|c|c|c}
        \toprule
        Test & \makecell{Recon. PSNR with\\ $\theta_0$ (dB)} & \makecell{Recon. PSNR with\\ estimated $\hat\theta_1$ (dB)} & Error traj. $\|\xi_1-\hat \xi_1\|_\infty$ & PSNR $\hat s_1$ (dB) \\
        \midrule
        1  &  17.26 &  33.04 & 0.039 & 44.15 \\
        2  &  13.09 &  29.42 & 0.045 & 36.30 \\
        3  &  13.85 &  37.69 & 0.032 & 47.91 \\
        4  &  18.50 &  34.95 & 0.026 & 40.26 \\
        5  &  15.46 &  33.65 & 0.050 & 44.35 \\
        6  &  20.57 &  31.43 & 0.008 & 58.20 \\
        7  &  20.90 &  33.92 & 0.021 & 44.89 \\
        8  &  18.60 &  34.29 & 0.031 & 49.30 \\
        9  &  10.88 &  33.05 & 0.016 & 41.56 \\
        \midrule
        Avg &  16.57 & 33.49 & 0.030 & 45.21 \\
        \bottomrule
    \end{tabular}
    \caption{\rev{Additional experiments for self-calibrated MRI with different images. The initial error $\|\xi_0-\xi_1\|_\infty$ on the trajectories is $5$ pixels for all test cases. We recall that $\hat s_1$ coincides with the estimated sensitivity maps and $\xi_1$ with the sampling trajectory.\label{tab:blind_MRI}}}
\end{table*}
}

\subsubsection{Computerized tomography}

\begin{figure*}[h]
    \def\subfigwidth{0.24\textwidth}
    \def\trimzoomleft{.3}
    \def\trimzoomlower{.48}
    \def\trimzoomright{.53}
    \def\trimzoomupper{.35}
    \centering
    \footnotesize
    \begin{tabular}{@{}cc@{}c@{}c@{}c@{}}
         & &
        \makecell{avg. err. $\alpha = 0.7^\circ$ \\ avg. err. $s = 1$ pixel} &
        \makecell{
        \begin{tikzpicture}
            \node[inner sep=0pt] (img) at (0,0) {
                \includegraphics[width=\subfigwidth]{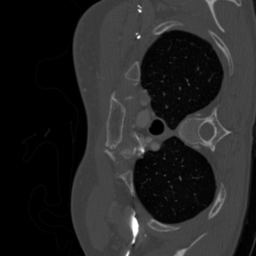}%
                \llap{\adjincludegraphics[width=0.1\textwidth,trim={{\trimzoomleft\width} {\trimzoomlower\height} {\trimzoomright\width} {\trimzoomupper\height}},clip,cfbox=red 0.5pt 0pt]{fig6_8_00.png}}
            };
            \draw [stealth-,red] (-0.35,0.1) -- (0.27,-0.25);
        \end{tikzpicture}
        \\ Ground truth} &
        \makecell{\includegraphics[width=\subfigwidth]{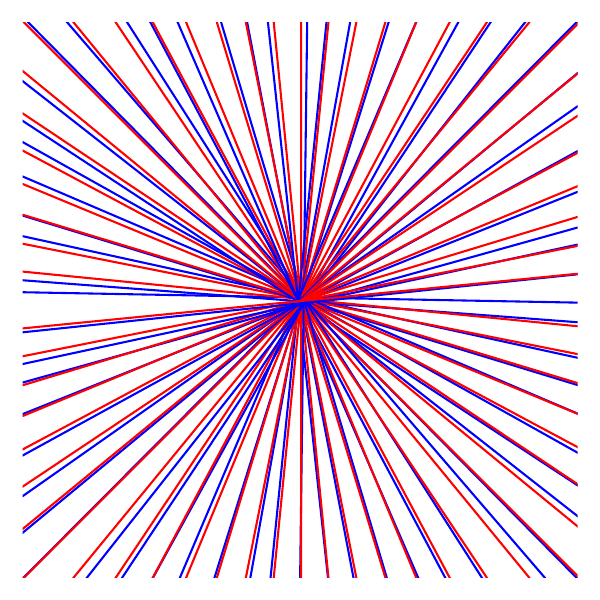}\\ \blue{$\theta_0$} vs \red{$\theta_1$}} \\
        \midrule

        \rotatebox[origin=c]{90}{Trained on $\thetaz_0$} &
        \makecell{
        \begin{tikzpicture}
            \node[inner sep=0pt] (img) at (0,0) {
                \includegraphics[width=\subfigwidth]{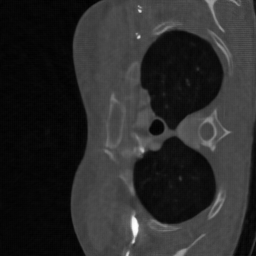}%
                \llap{\adjincludegraphics[width=0.1\textwidth,trim={{\trimzoomleft\width} {\trimzoomlower\height} {\trimzoomright\width} {\trimzoomupper\height}},clip,cfbox=red 0.5pt 0pt]{fig6_8_11_36.62.png}}
            };
            \draw [stealth-,red] (-0.35,0.1) -- (0.27,-0.25);
        \end{tikzpicture}
        \\ $36.62$dB} &
        \makecell{
        \begin{tikzpicture}
            \node[inner sep=0pt] (img) at (0,0) {
                \includegraphics[width=\subfigwidth]{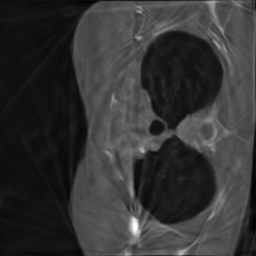}%
                \llap{\adjincludegraphics[width=0.1\textwidth,trim={{\trimzoomleft\width} {\trimzoomlower\height} {\trimzoomright\width} {\trimzoomupper\height}},clip,cfbox=red 0.5pt 0pt]{fig6_8_12_29.37.png}}
            };
            \draw [stealth-,red] (-0.35,0.1) -- (0.27,-0.25);
        \end{tikzpicture}
        \\ $29.37$dB} &
        \makecell{
        \begin{tikzpicture}
            \node[inner sep=0pt] (img) at (0,0) {
                \includegraphics[width=\subfigwidth]{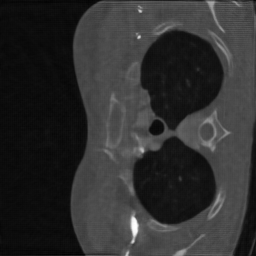}%
                \llap{\adjincludegraphics[width=0.1\textwidth,trim={{\trimzoomleft\width} {\trimzoomlower\height} {\trimzoomright\width} {\trimzoomupper\height}},clip,cfbox=red 0.5pt 0pt]{fig6_8_13_34.07.png}}
            };
            \draw [stealth-,red] (-0.35,0.1) -- (0.27,-0.25);
        \end{tikzpicture}
        \\ $34.07$dB} &
        \makecell{\includegraphics[width=\subfigwidth]{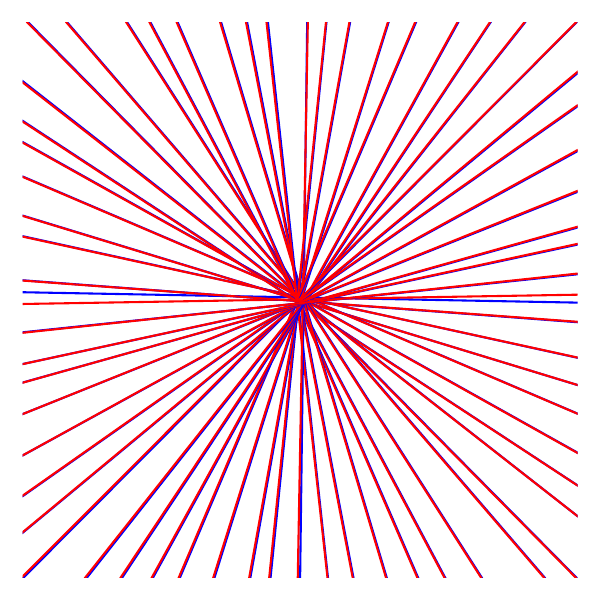}\\
        avg. err. $\hat\alpha_1=0.11^\circ$\\
        avg. err. $\hat s_1=0.3$ pixel} \\

        \rotatebox[origin=c]{90}{Trained on a family} &
        \makecell{
        \begin{tikzpicture}
            \node[inner sep=0pt] (img) at (0,0) {
                \includegraphics[width=\subfigwidth]{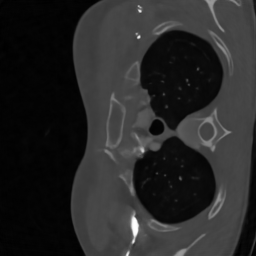}%
                \llap{\adjincludegraphics[width=0.1\textwidth,trim={{\trimzoomleft\width} {\trimzoomlower\height} {\trimzoomright\width} {\trimzoomupper\height}},clip,cfbox=red 0.5pt 0pt]{fig6_8_21_39.54.png}}
            };
            \draw [stealth-,red] (-0.35,0.1) -- (0.27,-0.25);
        \end{tikzpicture}
        \\ $39.54$dB} &
        \makecell{
        \begin{tikzpicture}
            \node[inner sep=0pt] (img) at (0,0) {
                \includegraphics[width=\subfigwidth]{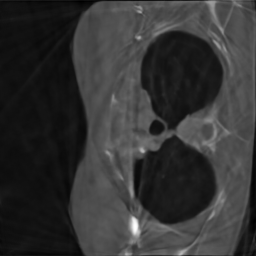}%
                \llap{\adjincludegraphics[width=0.1\textwidth,trim={{\trimzoomleft\width} {\trimzoomlower\height} {\trimzoomright\width} {\trimzoomupper\height}},clip,cfbox=red 0.5pt 0pt]{fig6_8_22_29.68.png}}
            };
            \draw [stealth-,red] (-0.35,0.1) -- (0.27,-0.25);
        \end{tikzpicture}
        \\ $29.68$dB} &
        \makecell{
        \begin{tikzpicture}
            \node[inner sep=0pt] (img) at (0,0) {
                \includegraphics[width=\subfigwidth]{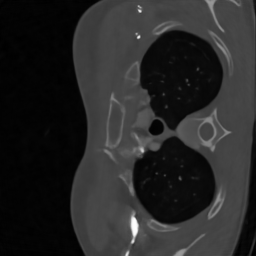}%
                \llap{\adjincludegraphics[width=0.1\textwidth,trim={{\trimzoomleft\width} {\trimzoomlower\height} {\trimzoomright\width} {\trimzoomupper\height}},clip,cfbox=red 0.5pt 0pt]{fig6_8_23_37.21.png}}
            };
            \draw [stealth-,red] (-0.35,0.1) -- (0.27,-0.25);
        \end{tikzpicture}
        \\ $37.21$dB} &
        \makecell{\includegraphics[width=\subfigwidth]{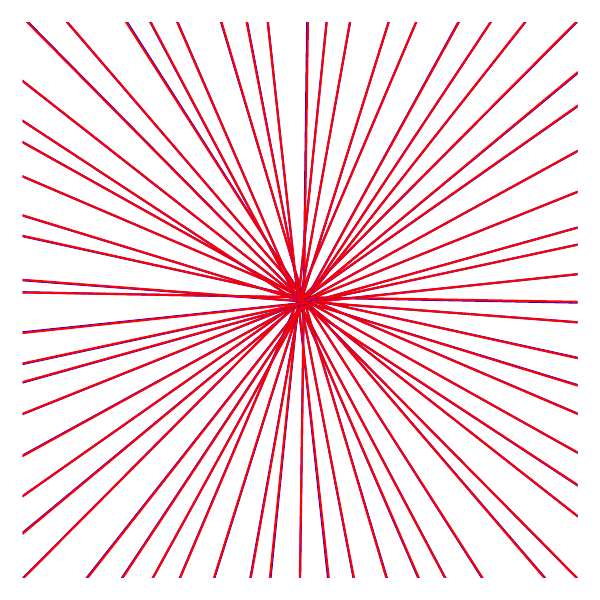}\\
        avg. err. $\hat\alpha_1=0.06^\circ$ \\
        avg. err. $\hat s_1=0.08$ pixel} \\

        \rotatebox[origin=c]{90}{P\&P} &
        \makecell{
        \begin{tikzpicture}
            \node[inner sep=0pt] (img) at (0,0) {
                \includegraphics[width=\subfigwidth]{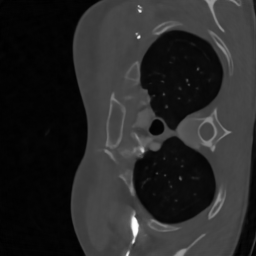}%
                \llap{\adjincludegraphics[width=0.1\textwidth,trim={{\trimzoomleft\width} {\trimzoomlower\height} {\trimzoomright\width} {\trimzoomupper\height}},clip,cfbox=red 0.5pt 0pt]{fig6_8_31_39.53.png}}
            };
            \draw [stealth-,red] (-0.35,0.1) -- (0.27,-0.25);
        \end{tikzpicture}
        \\ $39.53$dB} &
        \makecell{
        \begin{tikzpicture}
            \node[inner sep=0pt] (img) at (0,0) {
                \includegraphics[width=\subfigwidth]{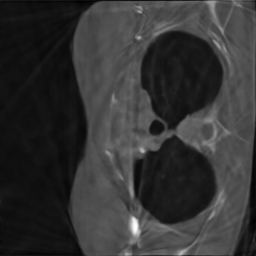}%
                \llap{\adjincludegraphics[width=0.1\textwidth,trim={{\trimzoomleft\width} {\trimzoomlower\height} {\trimzoomright\width} {\trimzoomupper\height}},clip,cfbox=red 0.5pt 0pt]{fig6_8_32_29.68.png}}
            };
            \draw [stealth-,red] (-0.35,0.1) -- (0.27,-0.25);
        \end{tikzpicture}
        \\ $29.68$dB} &
        \makecell{
        \begin{tikzpicture}
            \node[inner sep=0pt] (img) at (0,0) {
                \includegraphics[width=\subfigwidth]{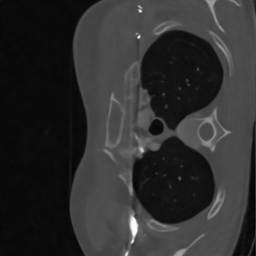}%
                \llap{\adjincludegraphics[width=0.1\textwidth,trim={{\trimzoomleft\width} {\trimzoomlower\height} {\trimzoomright\width} {\trimzoomupper\height}},clip,cfbox=red 0.5pt 0pt]{fig6_8_33_33.48.png}}
            };
            \draw [stealth-,red] (-0.35,0.1) -- (0.27,-0.25);
        \end{tikzpicture}
        \\ $33.48$dB} &
        \makecell{\includegraphics[width=\subfigwidth]{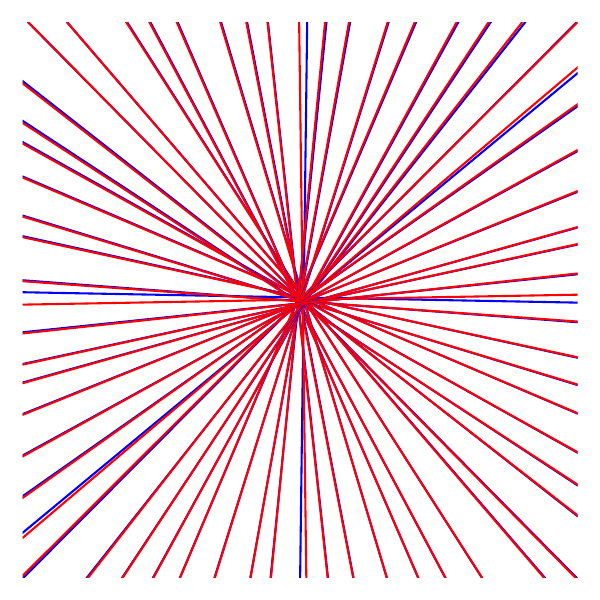}\\
        avg. err. $\hat\alpha_1=0.17^\circ$\\
        avg. err. $\hat s_1=0.13$ pixel} \\

         &
        $\Nc^a(\wz, \thetaz_1, \yz)$ &
        $\Nc^a(\wz, \thetaz_0, \yz)$ &
        $\Nc^a(\wz, \hat\thetaz_1, \yz)$ &
        Lines \\
    \end{tabular}
    \caption{\rev{Self-calibrated computerized tomography. 
    \emph{1st column:} reconstruction with a perfect knowledge of the forward model $\thetaz_1=(\alpha_1,s_1)$.
    \emph{2nd column:} reconstruction assuming the wrong forward model $\thetaz_0$. 
    \emph{3rd column:} reconstruction using the estimated forward model $\hat\thetaz_1$. 
    \emph{4th column:} true $\thetaz_1$ (blue) and estimated $\hat\thetaz_1$ parameters (red) of the forward model. 
    We display the average angle error and the average shift error.
    \emph{2nd row:} trained on $\thetaz_0$. 
    \emph{3rd row:} trained on a family of operators. 
    \emph{4th row:} using a P\&P prior. 
    The PSNR of the reconstructed image are indicated below each image.
    The noise level given to the P\&P denoiser has been tuned as to yield the best PSNR on the non-blind problem. Other models do not require tuning at evaluation.}}
    \label{tab:blind:ct}
\end{figure*}
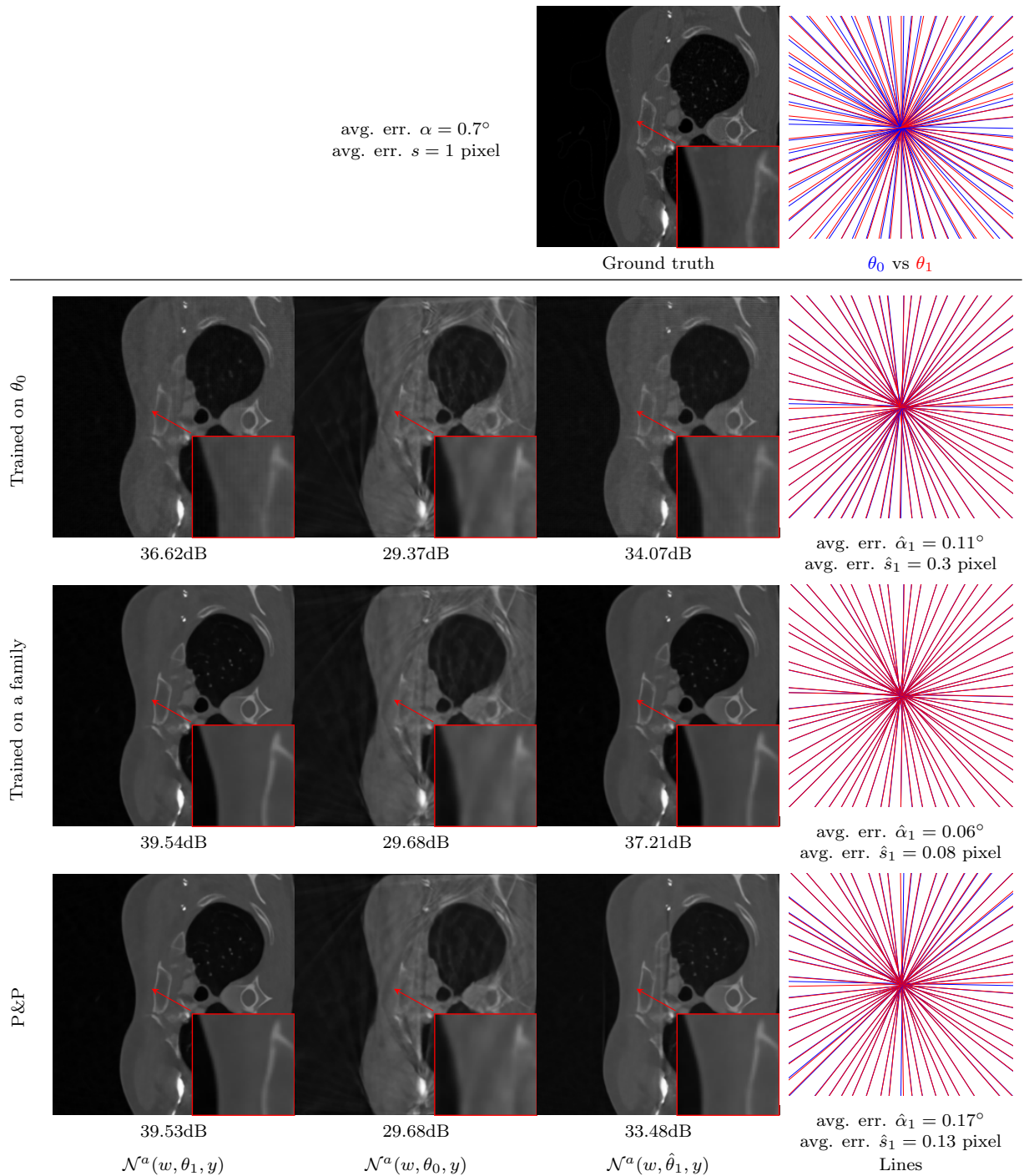

\rev{
In this application, a model mismatch might occur due to the motion of a patient in the scanner. 
Correcting this mismatch is essential.  
Not accounting for it, can result in severe artifacts including some details loss and blur as can be seen in Fig. \ref{tab:blind:ct}.}

\rev{
To identify the forward model, we ran the Adam optimizer on the parameters $\theta = (\alpha,s)$ for $2000$ iterations. 
In this application, $\alpha$ represents the angle of the parallel shots and $s$ their shift at origin.
All the reconstruction methods are able to significantly reduce the model mismatch, passing from maximal angles shifts of 7 degrees to less that 1 degree. 
Similarly, the shifts at origin are reduced from more than a pixel to about $0.3$ pixel. 
The reconstruction performance is significantly improved after estimating the forward model with PSNR increases of $4$dB and more.
The neural network trained on a family provides the best reconstruction results on this example. 
}

\rev{
Similarly to blind MRI, Table \ref{tab:blind_CT} shows that the ``deep unrolled prior'' method consistently provides good estimates of the forward model and significantly improves the reconstruction quality for the CT experiments.
\begin{table*}
    \footnotesize
    \centering
    \begin{tabular}{c|c|c|c|c}
        \toprule
        Test & \makecell{Recon. PSNR with\\ $\theta_0$ (dB)} & \makecell{Recon. PSNR with\\ estimated $\hat\theta_1$ (dB)} & Shift err. $\|s_1-\hat s_1\|_\infty$ & Angle err. $\|\alpha_1-\hat \alpha_1\|_\infty$ \\
        \midrule
        1  &  28.16 & 33.91 & 0.52 & 0.25 \\
        2  &  30.63 & 38.17 & 0.66 & 0.31 \\
        3  &  27.10 & 32.47 & 0.72 & 0.33 \\
        4  &  26.93 & 34.09 & 0.36 & 0.28 \\
        5  &  26.81 & 34.26 & 0.42 & 0.22 \\
        6  &  27.52 & 37.89 & 0.62 & 0.12 \\
        7  &  29.68 & 37.21 & 0.57 & 0.34 \\
        8  &  27.44 & 37.16 & 0.68 & 0.40 \\
        9  &  26.89 & 35.84 & 0.92 & 0.38 \\
        \midrule
        Avg & 27.91 & 35.67 & 0.61 & 0.29 \\
        \bottomrule
    \end{tabular}
    \caption{\rev{Additional experiments for self-calibrated CT with different images and operators.
    For all test cases, the initial angle error is $\|\alpha_0-\alpha_1\|_\infty=1.3^\circ$ and $\|s_0-s_1\|_\infty=1$ pixel.
    We see that the ``deep unrolled prior'' method provides good estimates of the true parameters $\theta_1=(\alpha_1, s_1)$ in all test cases. \label{tab:blind_CT}}}
\end{table*}
}

\subsubsection{Blind deblurring}

\begin{figure*}[h]
    \def\subfigwidth{0.2\textwidth}
    \def\trimzoomleft{.1}
    \def\trimzoomright{.7}
    \def\trimzoomlower{.7}
    \def\trimzoomupper{.1}
    \centering
    \footnotesize
    \begin{tabular}{cccc}
        \makecell{
        \begin{tikzpicture}
            \node[inner sep=0pt] (img) at (0,0) {
                \includegraphics[width=\subfigwidth]{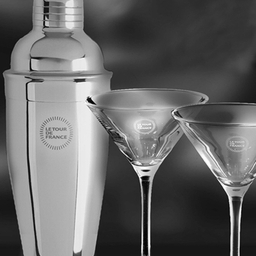}%
                \llap{\adjincludegraphics[width=0.1\textwidth,trim={{\trimzoomleft\width} {\trimzoomlower\height} {\trimzoomright\width} {\trimzoomupper\height}},clip,cfbox=red 0.5pt 0pt]{fig7_6_truex.png}}
            };
            \draw [stealth-,red] (-0.8,0.8) -- (-0.02,0.02);
        \end{tikzpicture}
        \\ Exact} &
        \makecell{
        \begin{tikzpicture}
            \node[inner sep=0pt] (img) at (0,0) {
                \includegraphics[width=\subfigwidth]{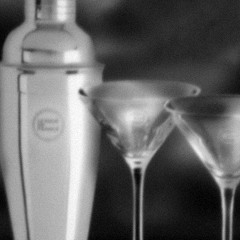}%
                \llap{\includegraphics[width=1cm,cfbox=green 0.5pt 0pt]{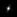}
                \adjincludegraphics[width=0.1\textwidth,trim={{\trimzoomleft\width} {\trimzoomlower\height} {\trimzoomright\width} {\trimzoomupper\height}},clip,cfbox=red 0.5pt 0pt]{fig7_6_y_25.41.png}}
            };
            \draw [stealth-,red] (-0.8,0.8) -- (-0.02,0.02);
        \end{tikzpicture}
        \\ Blurry $25.41$dB} &
        \makecell{
        \begin{tikzpicture}
            \node[inner sep=0pt] (img) at (0,0) {
                \includegraphics[width=\subfigwidth]{fig7_6_xideal_35.00.png}%
                \llap{
                \adjincludegraphics[width=0.1\textwidth,trim={{\trimzoomleft\width} {\trimzoomlower\height} {\trimzoomright\width} {\trimzoomupper\height}},clip,cfbox=red 0.5pt 0pt]{fig7_6_xideal_35.00.png}}
            };
            \draw [stealth-,red] (-0.8,0.8) -- (-0.02,0.02);
        \end{tikzpicture}
        \\ Non blind $35.00$dB} &
        \makecell{
        \begin{tikzpicture}
            \node[inner sep=0pt] (img) at (0,0) {
                \includegraphics[width=\subfigwidth]{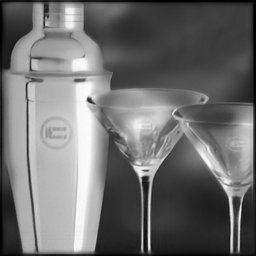}%
                \llap{\includegraphics[width=1cm,cfbox=green 0.5pt 0pt]{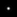}
                \adjincludegraphics[width=0.1\textwidth,trim={{\trimzoomleft\width} {\trimzoomlower\height} {\trimzoomright\width} {\trimzoomupper\height}},clip,cfbox=red 0.5pt 0pt]{fig7_6_xest_30.72.png}}
            };
            \draw [stealth-,red] (-0.8,0.8) -- (-0.02,0.02);
        \end{tikzpicture}
        \\ Blind $30.72$dB -- $26.93$dB}\\
        \makecell{
        \begin{tikzpicture}
            \node[inner sep=0pt] (img) at (0,0) {
                \includegraphics[width=\subfigwidth]{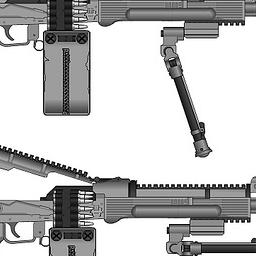}%
                \llap{\adjincludegraphics[width=0.1\textwidth,trim={{\trimzoomleft\width} {\trimzoomlower\height} {\trimzoomright\width} {\trimzoomupper\height}},clip,cfbox=red 0.5pt 0pt]{fig7_7_truex.png}}
            };
            \draw [stealth-,red] (-0.8,0.8) -- (-0.02,0.02);
        \end{tikzpicture}
        \\ Exact} &
        \makecell{
        \begin{tikzpicture}
            \node[inner sep=0pt] (img) at (0,0) {
                \includegraphics[width=\subfigwidth]{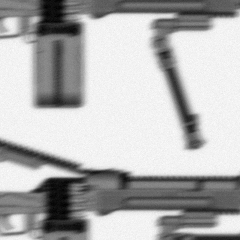}%
                \llap{\includegraphics[width=1cm,cfbox=green 0.5pt 0pt]{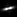}
                \adjincludegraphics[width=0.1\textwidth,trim={{\trimzoomleft\width} {\trimzoomlower\height} {\trimzoomright\width} {\trimzoomupper\height}},clip,cfbox=red 0.5pt 0pt]{fig7_7_y_18.11.png}}
            };
            \draw [stealth-,red] (-0.8,0.8) -- (-0.02,0.02);
        \end{tikzpicture}
        \\ Blurry $18.11$dB} &
        \makecell{
        \begin{tikzpicture}
            \node[inner sep=0pt] (img) at (0,0) {
                \includegraphics[width=\subfigwidth]{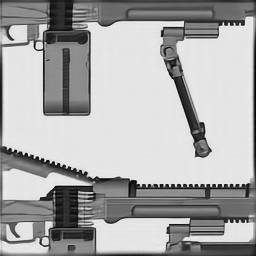}%
                \llap{
                \adjincludegraphics[width=0.1\textwidth,trim={{\trimzoomleft\width} {\trimzoomlower\height} {\trimzoomright\width} {\trimzoomupper\height}},clip,cfbox=red 0.5pt 0pt]{fig7_7_xideal_25.69.png}}
            };
            \draw [stealth-,red] (-0.8,0.8) -- (-0.02,0.02);
        \end{tikzpicture}
        \\ Non blind $25.69$dB} &
        \makecell{
        \begin{tikzpicture}
            \node[inner sep=0pt] (img) at (0,0) {
                \includegraphics[width=\subfigwidth]{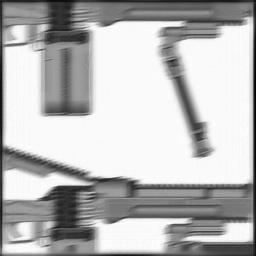}%
                \llap{\includegraphics[width=1cm,cfbox=green 0.5pt 0pt]{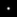}
                \adjincludegraphics[width=0.1\textwidth,trim={{\trimzoomleft\width} {\trimzoomlower\height} {\trimzoomright\width} {\trimzoomupper\height}},clip,cfbox=red 0.5pt 0pt]{fig7_7_xest_19.40.png}}
            };
            \draw [stealth-,red] (-0.8,0.8) -- (-0.02,0.02);
        \end{tikzpicture}
        \\ Blind $19.4$dB -- $9.16$dB}\\
        \makecell{
        \begin{tikzpicture}
            \node[inner sep=0pt] (img) at (0,0) {
                \includegraphics[width=\subfigwidth]{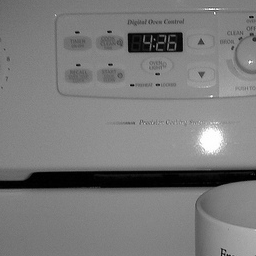}%
                \llap{\adjincludegraphics[width=0.1\textwidth,trim={{\trimzoomleft\width} {\trimzoomlower\height} {\trimzoomright\width} {\trimzoomupper\height}},clip,cfbox=red 0.5pt 0pt]{fig7_8_truex.png}}
            };
            \draw [stealth-,red] (-0.8,0.8) -- (-0.02,0.02);
        \end{tikzpicture}
        \\ Exact} &
        \makecell{
        \begin{tikzpicture}
            \node[inner sep=0pt] (img) at (0,0) {
                \includegraphics[width=\subfigwidth]{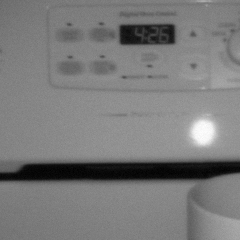}%
                \llap{\includegraphics[width=1cm,cfbox=green 0.5pt 0pt]{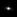}
                \adjincludegraphics[width=0.1\textwidth,trim={{\trimzoomleft\width} {\trimzoomlower\height} {\trimzoomright\width} {\trimzoomupper\height}},clip,cfbox=red 0.5pt 0pt]{fig7_8_y_31.55.png}}
            };
            \draw [stealth-,red] (-0.8,0.8) -- (-0.02,0.02);
        \end{tikzpicture}
        \\ Blurry $31.55$dB} &
        \makecell{
        \begin{tikzpicture}
            \node[inner sep=0pt] (img) at (0,0) {
                \includegraphics[width=\subfigwidth]{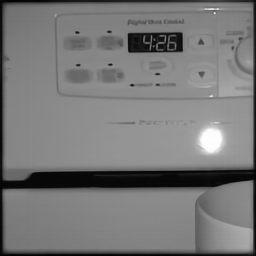}%
                \llap{
                \adjincludegraphics[width=0.1\textwidth,trim={{\trimzoomleft\width} {\trimzoomlower\height} {\trimzoomright\width} {\trimzoomupper\height}},clip,cfbox=red 0.5pt 0pt]{fig7_8_xideal_37.29.png}}
            };
            \draw [stealth-,red] (-0.8,0.8) -- (-0.02,0.02);
        \end{tikzpicture}
        \\ Non blind $37.29$dB} &
        \makecell{
        \begin{tikzpicture}
            \node[inner sep=0pt] (img) at (0,0) {
                \includegraphics[width=\subfigwidth]{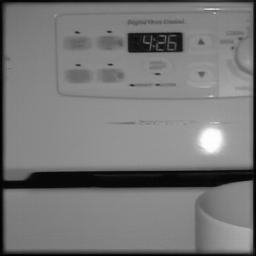}%
                \llap{\includegraphics[width=1cm,cfbox=green 0.5pt 0pt]{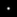}
                \adjincludegraphics[width=0.1\textwidth,trim={{\trimzoomleft\width} {\trimzoomlower\height} {\trimzoomright\width} {\trimzoomupper\height}},clip,cfbox=red 0.5pt 0pt]{fig7_8_xest_35.83.png}}
            };
            \draw [stealth-,red] (-0.8,0.8) -- (-0.02,0.02);
        \end{tikzpicture}
        \\ Blind $35.83$dB -- $26.20$dB}\\
        \makecell{
        \begin{tikzpicture}
            \node[inner sep=0pt] (img) at (0,0) {
                \includegraphics[width=\subfigwidth]{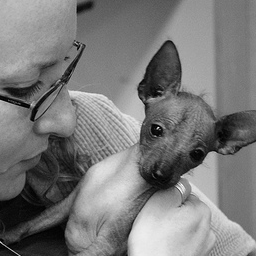}%
                \llap{\adjincludegraphics[width=0.1\textwidth,trim={{\trimzoomleft\width} {\trimzoomlower\height} {\trimzoomright\width} {\trimzoomupper\height}},clip,cfbox=red 0.5pt 0pt]{fig7_5_truex.png}}
            };
            \draw [stealth-,red] (-0.8,0.8) -- (-0.02,0.02);
        \end{tikzpicture}
        \\ Ground truth} &
        \makecell{
        \begin{tikzpicture}
            \node[inner sep=0pt] (img) at (0,0) {
                \includegraphics[width=\subfigwidth]{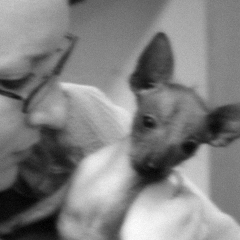}%
                \llap{\includegraphics[width=1cm,cfbox=green 0.5pt 0pt]{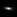}
                \adjincludegraphics[width=0.1\textwidth,trim={{\trimzoomleft\width} {\trimzoomlower\height} {\trimzoomright\width} {\trimzoomupper\height}},clip,cfbox=red 0.5pt 0pt]{fig7_5_y_27.26.png}}
            };
            \draw [stealth-,red] (-0.8,0.8) -- (-0.02,0.02);
        \end{tikzpicture}
        \\ Blurry $27.26$dB} &
        \makecell{
        \begin{tikzpicture}
            \node[inner sep=0pt] (img) at (0,0) {
                \includegraphics[width=\subfigwidth]{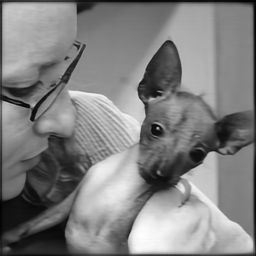}%
                \llap{
                \adjincludegraphics[width=0.1\textwidth,trim={{\trimzoomleft\width} {\trimzoomlower\height} {\trimzoomright\width} {\trimzoomupper\height}},clip,cfbox=red 0.5pt 0pt]{fig7_5_xideal_32.31.png}}
            };
            \draw [stealth-,red] (-0.8,0.8) -- (-0.02,0.02);
        \end{tikzpicture}
        \\ Non blind $32.31$dB} &
        \makecell{
        \begin{tikzpicture}
            \node[inner sep=0pt] (img) at (0,0) {
                \includegraphics[width=\subfigwidth]{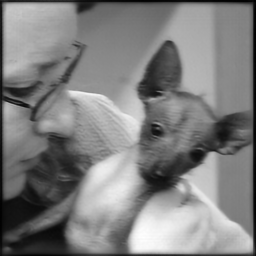}%
                \llap{\includegraphics[width=1cm,cfbox=green 0.5pt 0pt]{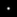}
                \adjincludegraphics[width=0.1\textwidth,trim={{\trimzoomleft\width} {\trimzoomlower\height} {\trimzoomright\width} {\trimzoomupper\height}},clip,cfbox=red 0.5pt 0pt]{fig7_5_xest_29.11.png}}
            };
            \draw [stealth-,red] (-0.8,0.8) -- (-0.02,0.02);
        \end{tikzpicture}
        \\ Blind $29.11$dB -- $19.02$dB}\\
        \makecell{
        \begin{tikzpicture}
            \node[inner sep=0pt] (img) at (0,0) {
                \includegraphics[width=\subfigwidth]{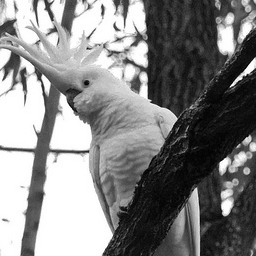}%
                \llap{\adjincludegraphics[width=0.1\textwidth,trim={{\trimzoomleft\width} {\trimzoomlower\height} {\trimzoomright\width} {\trimzoomupper\height}},clip,cfbox=red 0.5pt 0pt]{fig7_2_truex.png}}
            };
            \draw [stealth-,red] (-0.8,0.8) -- (-0.02,0.02);
        \end{tikzpicture}
        \\ Exact} &
        \makecell{
        \begin{tikzpicture}
            \node[inner sep=0pt] (img) at (0,0) {
                \includegraphics[width=\subfigwidth]{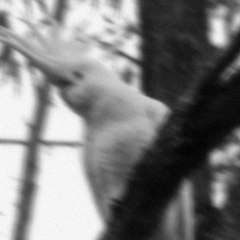}%
                \llap{\includegraphics[width=1cm,cfbox=green 0.5pt 0pt]{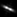}
                \adjincludegraphics[width=0.1\textwidth,trim={{\trimzoomleft\width} {\trimzoomlower\height} {\trimzoomright\width} {\trimzoomupper\height}},clip,cfbox=red 0.5pt 0pt]{fig7_2_y_19.64.png}}
            };
            \draw [stealth-,red] (-0.8,0.8) -- (-0.02,0.02);
        \end{tikzpicture}
        \\ Blurry $19.64$dB} &
        \makecell{
        \begin{tikzpicture}
            \node[inner sep=0pt] (img) at (0,0) {
                \includegraphics[width=\subfigwidth]{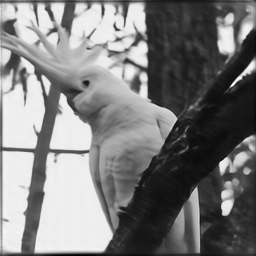}%
                \llap{
                \adjincludegraphics[width=0.1\textwidth,trim={{\trimzoomleft\width} {\trimzoomlower\height} {\trimzoomright\width} {\trimzoomupper\height}},clip,cfbox=red 0.5pt 0pt]{fig7_2_xideal_28.12.png}}
            };
            \draw [stealth-,red] (-0.8,0.8) -- (-0.02,0.02);
        \end{tikzpicture}
        \\ Non blind $28.12$dB} &
        \makecell{
        \begin{tikzpicture}
            \node[inner sep=0pt] (img) at (0,0) {
                \includegraphics[width=\subfigwidth]{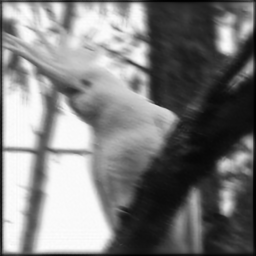}%
                \llap{\includegraphics[width=1cm,cfbox=green 0.5pt 0pt]{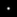}
                \adjincludegraphics[width=0.1\textwidth,trim={{\trimzoomleft\width} {\trimzoomlower\height} {\trimzoomright\width} {\trimzoomupper\height}},clip,cfbox=red 0.5pt 0pt]{fig7_2_xest_20.61.png}}
            };
            \draw [stealth-,red] (-0.8,0.8) -- (-0.02,0.02);
        \end{tikzpicture}
        \\ Blind $20.61$dB -- $9.03$dB}\\

    \end{tabular}
    \caption{\rev{The failure of deep unrolled prior for blind deblurring.
    \emph{1st column:} Ground truth image.
    \emph{2nd column:} Blurry image and the corresponding blur kernel. 
    \emph{3rd column:} Non blind reconstruction using the network trained on a family.
    \emph{4th column:} Blind reconstruction using the deep unrolled prior. The green box indicates the recovered kernel.
    The PSNR of the reconstructed image -- blur kernel are indicated below each image.
    From top to bottom: different images/blur kernels. As can be seen, the method always returns the same kernel, which is the smallest possible (an Airy pattern) in the family of Fresnel diffraction blurs. It therefore fails to estimate the operator.}}
    \label{tab:blind:blur}
\end{figure*}

\rev{
Finally, we present some results of the ``deep unrolled prior'' methodology in Fig. \ref{tab:blind:blur}. 
In this experiment, we simply used $3$ Zernike polynomials and optimized them globally using Bayesian optimization. 
Hence, the recovered kernels can be safely considered as the (near) global minimizers of the functional \ref{eq:blindinvmintheta}.
It appears that in every case, the method returns the same kernel, which is the one with the smallest possible extent in the family. 
It coincides with all Zernike coefficients being $0$, i.e. a Airy pattern. 

Hence, for this specific application, the deep unrolled prior methodology is not able to correctly identify the blur kernel. 
The reconstruction network still improves the image quality in average, but we cannot recommend this method for this application. 
Understanding the observed behavior requires further work, but shows that the proposed methodology does not work universally.  
}

\section{Conclusion}

In this work we proposed a training procedure to address the adaptivity and robustness issues in model-based unrolled  neural network for inverse problems.
We showed that a careful training leads to networks able to adapt to different forward operators without compromising the image quality. 
We also showed that minimizing a consistency term with the proposed networks makes it possible to solve challenging blind inverse problems in magnetic resonance imaging and computerized tomography.
\rev{In particular, we were able to correct trajectory errors and evaluate sensitivity maps convincingly for the first time.
The method can be seen as a new type of P\&P method to recover operators in blind inverse problems.
It however does not work for blind deblurring. This experiment shows that a theoretical analysis to better understand when and why the method works is needed.}

This work opens new interesting perspectives for computational imaging. 
A recent trend consists in optimizing the forward model and the reconstruction algorithm jointly (see e.g. \cite{weiss2021pilot,gossard2022spurious,wang2022b} for examples in MRI). With a reconstruction method capable of adapting to a vast family of operators, it becomes possible to restrict the attention to the optimization of the forward model only \cite{gossard2022bayesian}.

\rev{Extending the method to other applications seems relevant as well.}
An interesting perspective would be to add motion correction for MRI. 
A motion in the image domain translates to a phase modulation in the Fourier domain.
This is a critical issue in practice. 
The disconcerting ease with which we solved the estimation of trajectory shifts and sensitivity maps, sparks good hopes to solve this long resisting problem.

\appendix

\section{Detailed description of the forward models \label{app:precise_description}}

\paragraph{Parallel Magnetic Resonance Imaging}

Our aim here is to reconstruct images from under-sampled Fourier samples with unknown sensitivity maps associated to $J\in \N$ reception coils, and with inaccurate trajectories.
The parameter $\thetaz$ can be decomposed as $\thetaz=(\tau, \omega)$, where $\tau$ is a parameter describing the sensitivity maps and $\omega$ describes a perturbation of the sampling locations. 
To the best of our knowledge, these two problems have not been treated jointly in the literature yet.

Let $\Fc(\xi)$ denote the non-uniform Fourier transform (NUFT) \cite{potts2001fast} at frequencies $\xi=(\xi_1,\hdots, \xi_M)$, defined by 
\begin{equation*}
[\Fc(\xi)]_{m,n} = e^{-i\langle p_n, \xi_m\rangle}
\end{equation*}
where $(p_n)_{1\leq n\leq N}$ is a set of 2D positions on a grid. 
We construct a family of forward operators $\Ac = \left\{\Az(\xi, \thetaz), \xi\in \Xi, \thetaz\in \Theta \right\}$, where $\Xi\subset \R^{2\times M}$ is a set of 2D sampling schemes with $M$ sampling points. 
The parameter space $\Theta =  \Tc \times \Omega$ describes the set of admissible parameters for the sensitivity maps $\Tc$ and for the perturbation $\Omega$.
The measured signal $\yz=(\yz ^{(1)},\hdots, \yz^{(J)})$ is acquired through $J$ coils. 
The $m$-th measurement acquired by the $j$-th coil is defined by
\begin{equation}\label{eq:model_forward_MRI}
    y_m^{(j)} = \left[\Az(\xi, \thetaz)\xz\right]_{m,j} + \bz_{m,j} = \left[\Fc(h(\omega) \star\xi)\left(\xz\odot s(\tau^{(j)})\right)\right]_{m} + \bz_{m,j}.
\end{equation}
The mapping $s:\tau^{(j)}\in\R^T\mapsto s(\tau^{(j)})\in\C^N$ parametrizes the coil sensitivity maps.
Since the sensitivity maps are smooth, we use a parametrization based on thin plate splines \cite{duchon1977splines}. The total number of parameters that encode the sensitivity map is $T=104$.
It consists of the splines coefficients using $7\times 7$ regularly spaced control points, plus the coefficients of a first degree polynomial. This has to be multiplied by two for the real and imaginary parts.

Following \cite{vannesjo2016image,dietrich2016field}, we assume that the trajectory $\xi$ is perturbed by a convolution with an impulse response $h(\omega)$. The symbol $\star$ in \eqref{eq:model_forward_MRI} corresponds to a discrete convolution. Evaluating the convolution filter $h(\omega)$ is known as a challenging problem  that can be addressed with expensive field cameras \cite{dietrich2016field}. Here, in the spirit of \cite{vannesjo2016image}, we will rather treat it as a blind inverse problem. We parametrize $h$ as a linear combination of the form $h(\omega) = \sum_{o=1}^O \omega_o h_o$, where $(h_o)_{1\leq o\leq O}$ is an orthogonal basis. In practice, we simply use compactly supported filters of size $O=32$ and $(h_o)_{1\leq o\leq O}$ corresponds to the first $32$ elements of the canonical basis.

\paragraph{Computerized Tomography}

Our aim is to reconstruct images from parallel beam computerized tomography. 
The parameter $\thetaz$ describes the projection angles and shift at origin (allowing to model the patient motion).

We assume that the CT scan uses parallel beams and that it performs $J$ acquisitions with a receptor that has $M$ sensors, resulting in $J\times M$ measurements. 
In this application, the parameter $\thetaz = (\alpha,s)$ represents the angles $\alpha\in\R^J$ and the shifts at the origin $s\in\R^J$ that describe the beams trajectories.
If $m$ corresponds to the $m$-th pixel of the receptor and if we index the acquisitions by $1\leq j\leq J$, we get
\begin{equation}
    y_m^{(j)} = \iint_\Omega \xz(u_x,u_y)\, \delta_{u_x\cos(\alpha_j)+u_y\sin(\alpha_j)=p_m+s_j} \,\mathrm{d}u_x\mathrm{d}u_y + b_m^{(j)},
\end{equation}
with $\Omega=\left[-N_x/2,N_x/2\right]\times\left[-N_y/2,N_y/2\right]$ and $p \in \llbracket -M/2,\ldots,M/2-1\rrbracket$.
A perfect model would correspond to $\alpha$ being equispaced angles and $s=0$.
The forward model can be computed using the Fourier slice theorem. This corresponds to performing a 2D NUFT and we resort to the same library used for MRI (see \url{https://github.com/albangossard/Bindings-NUFFT-pytorch}).

\paragraph{Deblurring in optics}

In this application, we wish to solve problems appearing in \rev{diffraction limited systems}. 
The parameter $\thetaz$ describes the point spread function \rev{through the theory of diffraction.}
The acquisition model in this application simply reads
\begin{equation}
    \yz = h(\thetaz) \star \xz + \bz,
\end{equation}
where $h(\thetaz)$ is the blur kernel.
We consider blurs generated by Fresnel diffraction theory \cite{goodman1996introduction}.
\rev{This theory is the most commonly adopted in optics since the works of Zernike (see e.g. \cite{noll1976zernike,lakshminarayanan2011zernike}). It is widely used in microscopy or astronomy.}

The blur kernel is parameterized by a vector $\thetaz\in\R^7$ and the convolution kernel is expressed as
\begin{equation}\label{eq:zernike_parametrization}
    h(\thetaz) = c \left| \int_{\|\wz\|_2\leq f_c} \exp\left( 2i\pi \left[ \sum_{k=1}^{K} \theta_k Z_k + \langle u,\wz \rangle\right] \right)\,\mathrm{d}\wz \right|^2.
\end{equation}
In this expression, $f_c$ is a cutoff frequency and $c$ is a scaling parameter such that $\| h\|_1=1$.
The expansion $\sum_{k=1}^{K} \theta_k Z_k$ describes the pupil function of an objective with a circular aperture.
The functions $Z_k$ are Zernike polynomials and the vector $\thetaz$ parametrizes the so-called pupil function.

\section{Details on the pseudo-inverse and the resolvent}

The reconstruction networks all rely on the pseudo-inverse $A(\theta)^\dagger$ or on the resolvent $(A(\theta)^* A(\theta) + \lambda \Id)^{-1}$.
In all our experiments, we implemented them using a conjugate gradient algorithm run for a fixed number of iterations. In all cases, we set this number to ensure a relative residue below a threshold of $10^{-4}$.

\rev{The pseudo-inverse $\xz\mapsto\Az(\thetaz)^\dagger\xz$ in the reconstruction networks is approximated by solving the symmetric positive definite system $(A^*A+\epsilon \Id)x=y$. The parameter $\epsilon$ is set to a small value that was tuned manually for each application.}

\rev{For larger $\lambda$, the linear system is better conditioned and can be solved using less iterations with the same conjugate gradient iteration.}

\section*{Acknowledgments}
This work was supported by the ANR Micro-Blind (ANR-21-CE48-0008) and by the ANR LabEx CIMI (ANR-11-LABX-0040) within the French State Programme ``Investissements d’Avenir''.
P. Weiss acknowledges the support of AI Interdisciplinary Institute ANITI funding, through the French ``Investing for the Future— PIA3'' (ANR-19-PI3A-0004). This work was performed using HPC resources from GENCI-IDRIS (2021-AD011012210R1).
\rev{Parts of the code can be made available on demand.
The authors wish to thank Emmanuel Soubies, Valentin Debarnot, Nathanaël Munier, Frédéric de Gournay and the anonymous reviewers for their comments and advice which helped us improving the paper.
}

\clearpage
{\small
\bibliographystyle{plain}
\bibliography{./biblio}
}

\end{document}